\documentclass[10pt,journal,compsoc]{IEEEtran}

\ifCLASSOPTIONcompsoc
  \usepackage[nocompress]{cite}
\else
  \usepackage{cite}
\fi

\usepackage{microtype}
\usepackage[T1]{fontenc}
\usepackage[utf8]{inputenc}
\usepackage[french,main=english]{babel}
\usepackage{csquotes}
\usepackage[bookmarks=false]{hyperref} 
\usepackage{mathptmx}
\usepackage{amsmath}
\usepackage{amssymb}
\usepackage{amsfonts}
\usepackage{booktabs}  
\usepackage{multirow}
\usepackage{graphicx}
\usepackage{float}
\usepackage{graphbox}
\usepackage{caption}
\usepackage{subcaption}
\usepackage[ruled,vlined,linesnumbered]{algorithm2e}
\usepackage[export]{adjustbox}
\usepackage{threeparttable}
\usepackage{tabularx}

\newcommand\funcintext[1]{#1}

\usepackage{pifont}
\newcommand{\cmark}{\ding{51}}%
\newcommand{\xmark}{\ding{55}}

\usepackage{xcolor}
\definecolor{mygreen}{rgb}{0.55, 0.71, 0.0}
\definecolor{myblue}{rgb}{0.36, 0.54, 0.66}

\SetCommentSty{mycommfont}

\DeclareMathOperator*{\argmax}{argmax}
\DeclareMathOperator*{\concatenate}{concatenate}
\DeclareMathAlphabet{\mathcal}{OMS}{cmsy}{m}{n}

\newcommand\modelname{Vertical Attention Network}
\newcommand\modelacc{VAN}
\newcommand\tens[1]{#1}

\begin{document}
\sloppy
\title{End-to-end Handwritten Paragraph Text Recognition Using a \modelname{}}

% author names and IEEE memberships
% note positions of commas and nonbreaking spaces ( ~ ) LaTeX will not break
% a structure at a ~ so this keeps an author's name from being broken across
% two lines.
% use \thanks{} to gain access to the first footnote area
% a separate \thanks must be used for each paragraph as LaTeX2e's \thanks
% was not built to handle multiple paragraphs
%
%
%\IEEEcompsocitemizethanks is a special \thanks that produces the bulleted
% lists the Computer Society journals use for "first footnote" author
% affiliations. Use \IEEEcompsocthanksitem which works much like \item
% for each affiliation group. When not in compsoc mode,
% \IEEEcompsocitemizethanks becomes like \thanks and
% \IEEEcompsocthanksitem becomes a line break with idention. This
% facilitates dual compilation, although admittedly the differences in the
% desired content of \author between the different types of papers makes a
% one-size-fits-all approach a daunting prospect. For instance, compsoc 
% journal papers have the author affiliations above the "Manuscript
% received ..."  text while in non-compsoc journals this is reversed. Sigh.

\author{Denis~Coquenet,
        Clément~Chatelain,
        and~Thierry~Paquet% <-this % stops a space
\IEEEcompsocitemizethanks{
\IEEEcompsocthanksitem The authors are with the LITIS, France.\protect\\
E-mail: \{denis.coquenet,clement.chatelain,thierry.paquet\}@litislab.eu
\IEEEcompsocthanksitem D. Coquenet is with Rouen University and  Normandy University, France
\IEEEcompsocthanksitem C. Chatelain is with the INSA of Rouen, France
\IEEEcompsocthanksitem T. Paquet is with Rouen University, France
}% 
\thanks{Manuscript received ...; revised ...}}

% \markboth{Journal of \LaTeX\ Class Files,~Vol.~14, No.~8, August~2015}%
% {Shell \MakeLowercase{\textit{et al.}}: Bare Demo of IEEEtran.cls for Computer Society Journals}

\IEEEtitleabstractindextext{
\begin{abstract}
Unconstrained handwritten text recognition remains challenging for computer vision systems. Paragraph text recognition is traditionally achieved by two models: the first one for line segmentation and the second one for text line recognition. We propose a unified end-to-end model using hybrid attention to tackle this task. This model is designed to iteratively process a paragraph image line by line. It can be split into three modules. An encoder generates feature maps from the whole paragraph image. Then, an attention module recurrently generates a vertical weighted mask enabling to focus on the current text line features. This way, it performs a kind of implicit line segmentation. For each text line features, a decoder module recognizes the character sequence associated, leading to the recognition of a whole paragraph. We achieve state-of-the-art character error rate at paragraph level on three popular datasets: 1.91\% for RIMES, 4.45\% for IAM and 3.59\% for READ 2016. Our code and trained model weights are available at \url{https://github.com/FactoDeepLearning/VerticalAttentionOCR}.
\end{abstract}

}

\IEEEdisplaynontitleabstractindextext
\IEEEpeerreviewmaketitle

\maketitle

\IEEEraisesectionheading{\section{Introduction}\label{sec:introduction}}
\IEEEPARstart{O}{ffline} handwritten text recognition consists in recognizing the text in an image scanned from a document. An image of a word, a line or a paragraph of text, or even a full document is analyzed, and the sequence of characters that composes the text is expected as output. This paper focuses on paragraph text recognition based on an end-to-end segmentation-free neural network. 
Indeed, while line segmentation and handwriting recognition have been studied for decades now, they remain a challenging task. Moreover, they have rarely been studied and optimized together in one single trainable system. 

Historically, early works have applied segmentation at character level and each character was then classified. Later on, segmentation was applied at word level first, then at line level. But whatever the segmentation level considered, the problem related to the definition of the individual entities to be segmented remains ambiguous. In this study, we go a step further and process a paragraph of text without any explicit segmentation step, neither during training nor during decoding. However, pretraining on line images is used to improve the convergence and the recognition accuracy.

When switching from line images to paragraph images, one faces new challenges. Indeed, the model must include a way to locate the text regions (word or line for example) and to order them. This brings several new difficulties: 
\begin{itemize}
    \item The number of text regions varies from one paragraph to another.
    \item The layout diversity can be important due to multiple factors such as line spacing, horizontal alignment or slant.
    \item A reading order should be defined or learned in order to concatenate the recognized text regions together and obtain the final paragraph transcription.
\end{itemize}

To carry out the recognition of whole page images, current approaches still rely on a two-step approach. In a first step, the document is segmented into text regions, which are then recognized in a second step. The recognition is performed applying an optical model, thus the name Optical Character Recognition (OCR). 
Taken separately, each of these two steps brings rather good results but if considered together they show three major drawbacks. Firstly, they require ground truth segmentation labels as well as transcription labels at line level, which is very costly to produce by hand. Secondly, this two-step approach accumulates the errors of each individual step: segmentation errors induce OCR errors, while the OCR stage produces its own errors. And finally, a two-step strategy implies that any modification of one stage should lead to retraining both stages so as to optimally combine both stages in the whole model. In addition, using a prior explicit segmentation step raises the question of the definition of a line. Baseline, X-height or bounding boxes are examples of target labels for segmentation that have been frequently used in the literature, all with their pros and cons \cite{Renton2018}. It is also frequent to find variability in labels from one annotator to another. We can also highlight that the reading order is defined by hand, based on the coordinates of the text regions; this could lead to some errors in the case of rather slanted lines.

This paper aims at providing a model freed from all of these constraints. We suggest using a segmentation-free model that processes whole handwritten paragraphs using an attention process. In this model, character recognition and implicit line segmentation are learned in an end-to-end fashion, so as to optimize both processes altogether. Most of the contributions of the literature have successfully used neural networks for line segmentation and text line recognition as well, reaching state-of-the-art results. In addition, Attention Neural Networks have been successfully applied for many other tasks such as translation \cite{Bahdanau2014}, speech recognition \cite{Chorowski2015}, image captioning \cite{Xu2015} or even OCR applied at line level \cite{Chowdhurry2018}. This leads us to think that both tasks (segmentation and recognition) could be handled by a single neural network with attention as a control block. 

In this paper, we propose an encoder-decoder architecture using a hybrid (content-based and location-based) attention mechanism to process whole paragraph images. The idea is to recurrently recognize the lines so as to remain in a one-dimensional sequence alignment problem between the recognized text lines and their ground truth transcription. This alignment is achieved during training using the standard Connectionist Temporal Classification (CTC) loss \cite{CTC}. To this end, we first use an encoder to extract two-dimensional features from the input paragraph images. These features account for the characters while preserving their location. The text line location and their reading order are both handled and learned by the attention module. This attention mechanism aims at focusing on specific feature rows to generate text line representations. They are generated recurrently, one at a time, from the first one to the last one, through the attention mechanism. The decoder recognizes a sequence of characters from each text line representation. A whitespace character is inserted between each recognized text line  to get the final paragraph.

In brief, we make the following contributions:
\begin{itemize}
    \item We propose the \modelname{}: a novel encoder-decoder architecture using hybrid attention for text recognition at paragraph level.
    
    \item The approach relies on an implicit line segmentation performed in the latent space of a deep model.
    
    \item We achieve state-of-the-art results on RIMES, IAM and READ 2016 datasets compared to paragraph-level approaches.
    
    \item We compare favorably this architecture with a standard two-step approach based on line segmentation followed by character recognition.

\end{itemize}

This paper is organized as follows. Related works are presented in Section \ref{section-related-works}. Section \ref{section-archi} is dedicated to the presentation of the proposed architecture. Section \ref{section-experiment-cond} is devoted to the experimental environment. It provides, \textit{inter alia}, description of datasets, training and implementation details. Experiments and results are detailed in Section \ref{section-experiments}. We carried out extensive experiments in Section \ref{section-additional-exp}. Section \ref{section-discussion} provides a discussion of the model and we draw conclusions in Section \ref{section-conclusion}.
\section{Related Works}
\label{section-related-works}

In the literature, only very few works have been devoted to multi-line text recognition, and most studies have concentrated on isolated lines recognition. We can classify these pioneer works into two categories: those using an explicit word/line segmentation, requiring segmentation and transcription labels, and those without any segmentation, only requiring transcription labels.

\subsection{Approaches using explicit line segmentation}

Explicit line segmentation approaches are two-step methods that sequentially detect the text lines of a paragraph, and then proceed to their recognition.
To our knowledge, \cite{Chung2020} is the only reference that gathers in the same study segmentation and text line recognition as two separate networks. However, a lot of works focus on one of these two tasks separately.

Early segmentation techniques \cite{Sulem2007} can be classified in three categories as suggested in \cite{Papavassiliou2011}. First, the projection-based methods, which consider the boundaries between lines as valleys of vertical projection profile \cite{Weliwitage2005}. Second, the grouping methods; it consists in grouping rows of connected components according to heuristic rules \cite{Feldbach2001}. The last category is the smearing methods which use blurring filters combined with binarization or active contours for example \cite{Swaileh2015}.  It is now generally handled by a Fully Convolutional Network (FCN) as in \cite{Renton2018, DHSegment, ARUNet}. 

Regarding handwritten lines recognition, it was first solved using handcrafted features and Hidden Markov Model (HMM) \cite{Ploetz2009,Yacoubi1999}. However, those models lacked discriminative power and were limited when dealing with long term dependencies in sequences. Hybrid systems combining HMM with Neural Networks (NN) were proposed, leading to better results over standard HMM: HMM with MultiLayer Perceptron (HMM+MLP) \cite{Bengio1995,Gilloux1995,Salicetti1996,Knerr1998}, HMM with Convolutional Neural Network (HMM+CNN) \cite{Bluche2013} or HMM with Recurrent Neural Network (HMM+RNN) \cite{Senior1995,Frinken2009}. Currently, the state of the art is reached with neural networks.

Many kinds of architectures have been proposed: Multi-Dimensional Long-Short Term Memory (MDLSTM) \cite{Voigtlaender2016}, hybrid CNN and Bidirectional LSTM (BLSTM) \cite{Wigington2017,Bruno2016} and more recently encoder-decoder with attention \cite{Michael2019}, Gated CCN (GCNN) \cite{Coquenet2019} and Gated FCN (GFCN) \cite{Yousef_line,Coquenet2020}.
Except for attention-based character recognition models, which use cross-entropy loss, training an OCR model at line level was made possible thanks to the CTC. Indeed, it enables to align output sequences of characters with input sequences of features (or pixels) of different and variable lengths.

Recently, one can notice a trend toward gathering segmentation and recognition together. For the segmentation stage, we can distinguish two approaches: the first one comes from object detection methods and the second one is based on start-of-line prediction.

The models proposed in \cite{Carbonell2019,Carbonell2020,Chung2020} follow the object detection approach: they are based on word (or line) bounding boxes prediction. They use a Region Proposal Network (RPN) combined with a non-maximal suppression process and Region Of Interest (ROI) pooling to obtain bounding boxes for each word or line in the input image. OCR is then applied on those boxes to recognize the text. In \cite{Carbonell2019,Carbonell2020}, the authors propose such a model and introduce a multi-task end-to-end architecture working at word level. 
In \cite{Chung2020}, the focus is on the modular approach. The authors propose a pipeline with four modules: a passage identification finds the handwritten text areas, then an object-detection-based word-level segmentation is applied and words are merged into lines afterwards. Finally, OCR is used on those lines, combined with a language model.

Other works focus on predicting start-of-line coordinates and heights.
In \cite{Moysset2017}, a CNN+MDLSTM is used to predict start-of-line references and an MDLSTM is used for text lines recognition with a dedicated end-of-line token. This is particularly useful in the context of multi-column texts. In \cite{Wigington2018,Wigington2019}, a VGG-11-based CNN is used as start-of-line predictor. Then, a recurrent process predicts the next position based on the current one until the end of the line, generating a normalized line. A CNN+BLSTM is finally used as OCR on lines. Some examples with start-of-line, line segmentation and line transcription labels are needed to pretrain the different subnetworks individually; the network is then able to work only with transcription labels. The approach proposed in \cite{Wigington2019} is similar to the one of \cite{Wigington2018}, but it can handle transcriptions without line breaks.

\subsection{Segmentation-free approaches}
Unlike explicit line segmentation approaches, segmentation-free approaches output the transcription result of a whole paragraph, performing recognition without a prior segmentation step.
Among the segmentation-free approaches, one can notice two trends: a first one based on recurrent attention mechanisms and a second one exploiting the two-dimensional nature of the task, based on a one-step process.

To our knowledge, \cite{Bluche2016,Bluche2017} are the only works applying attention mechanisms to the task of multi-line text recognition, achieving a kind of implicit line/character segmentation. These works propose an encoder-decoder architecture with attention blocks at line and character levels, respectively. Both architectures use a CNN+MDLSTM encoder and an MDLSTM-based attention module. The encoder produces feature maps from the input image while the attention module recurrently generates line or character representations applying a weighted sum between the attention weights and the features. Finally, the decoder outputs character probabilities from this representation. These architectures require pretraining on line-level images, but they do not need line breaks to be included in the transcription labels.

Only two other works present segmentation-free approaches, focusing on the dimensional aspect to recognize the text without a recurrent process. They are \cite{Schall2018} and \cite{Yousef2020}.
In \cite{Schall2018}, the authors propose a two-dimensional version of the CTC: the Multi-Dimensional Connectionist Classification (MDCC). Using a Conditional Random Field (CRF), ground truth transcription sequences are converted into a two-dimensional model (a 2D CRF) able to represent multiple lines. A line separator label is introduced in addition to the CTC null symbol (also called blank label). The CRF graph enables to jump from one line to the following one, whatever the position in the current line. As for the CTC, repeating labels account for only one. An MDLSTM-based network is used to generate probabilities in two dimensions, preserving the spatial nature of the input image. Pretraining is also used beforehand, this time at word level, with the standard CTC.

Finally, in \cite{Yousef2020} the authors focus on learning to unfold the input paragraph image \textit{i.e.} into a single text line. The system is trained to concatenate text lines to obtain a single large line before character recognition takes place. This is mainly carried out with bi-linear interpolation layers combined with an FCN encoder. This transformation network enables to use the standard CTC loss and to process the image in a single step. Moreover, it neither needs pretraining nor line breaks in the transcriptions and achieves state-of-the-art results.

The approaches proposed in \cite{Bluche2017} and \cite{Schall2018} are the first attempts reported in the literature towards end-to-end handwritten paragraph recognition, but they have remained below the state-of-the-art results on the IAM dataset \cite{IAM}. The interesting segmentation-free approaches presented in \cite{Bluche2016} and \cite{Yousef2020} reached competitive results.

The recurrent process used in \cite{Bluche2016} enables to model dependencies between lines, but is computationally expensive due to the MDLSTM layers, which could lead to high training and prediction times. The fully convolutional model involved in \cite{Yousef2020} enables high computation parallelization, but its recurrence-free process does not enable to model dependencies between lines. In addition, it implies a large number of parameters, of around 16.4 million. 

In this work, we suggest combining the advantages of both approaches to design a fast, efficient and lightweight end-to-end model for paragraph recognition.
The proposed model is based on an encoder-decoder architecture using attention as in \cite{Bluche2016}, but we use an FCN encoder and an attention module without recurrent layers to reduce the computation time while implying few parameters at the same time. 

\section{Architecture}
\label{section-archi}

We propose an end-to-end model, called \modelname{} (\modelacc{}), following the encoder-decoder principles with an attention module. It takes as input an image $X$ and recurrently recognizes the $L$ text lines $[y_1, ..., y_L]$ present in it. Each text line $y_t$ is a sequence of tokens from an alphabet $A$, with $|A|=N$.

\begin{figure*}[htbp!]
\centering
\includegraphics[width=\textwidth]{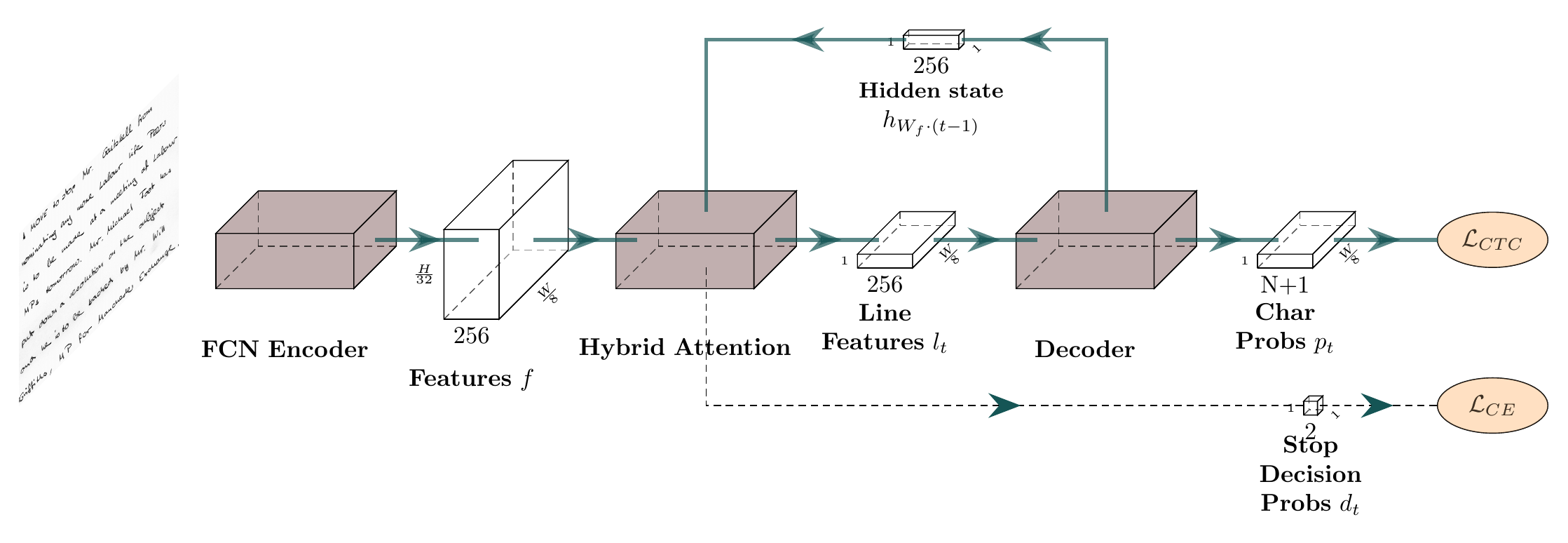}
        \caption{Architecture overview.}
        \label{fig:overview}
\end{figure*}

The overall model is presented in Figure \ref{fig:overview}. It first extracts features $f$ from $X$. We wanted the encoder stage to be modular enough in order to be plugged in different architectures dedicated to text lines, paragraphs or documents recognition without needing any adaptation. To this end, we chose a Fully Convolutional Network encoder that can deal with input images of variable heights and widths, possibly containing multiple lines of text. The attention module is the main control block: it recurrently produces vertical attention weights that focus where to select the next features for the recognition of the next text line. This way, it recurrently generates as many text line features/representations $l_t$ as there are text lines in the input image. It also detects the end of the paragraph leading to the end of the whole process. Finally, the decoder stage produces character probabilities $p_t$ for each frame of each generated line features, and the best path decoding strategy is used. The model is trained using a combination of two losses: the CTC loss for the recognition task, through the line-by-line alignment between recognized text lines (through its probabilities lattice $p_t$) and ground truth line transcriptions $y_t$, and the cross-entropy loss for the end-of-paragraph detection. We now describe each module in detail.

\subsection{Encoder}
An FCN encoder is used to extract features from the input paragraph images. It takes as input an image $\tens{X} \in \mathbb{R}^{H \times W \times C}$, H, W and C being respectively the height, the width and the number of channels (C=1 for a grayscale image, C=3 for a RGB image). Then, it outputs feature maps for the whole paragraph image: $\tens{f} \in \mathbb{R}^{H_f \times W_f \times C_f}$ with $H_f=\frac{H}{32}$, $W_f = \frac{W}{8}$ and $C_f = 256$. Those features must contain enough information to recognize the characters afterwards, while preserving the two-dimensional nature of the task.

\begin{figure*}[h!]
\centering
    \includegraphics[width=\textwidth]{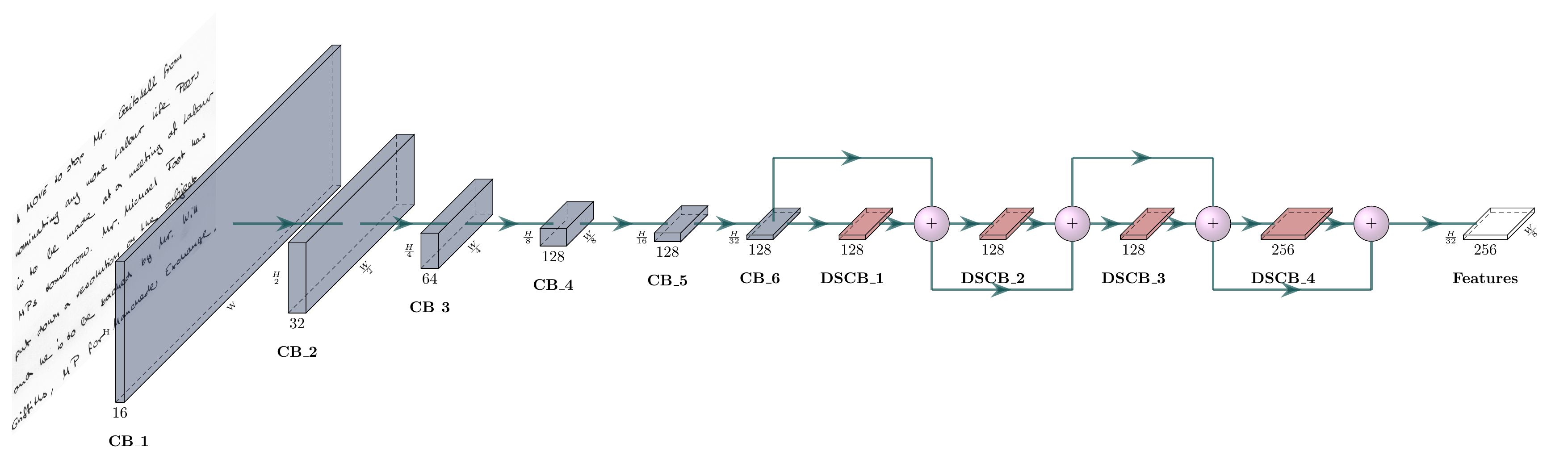}
        \caption{FCN Encoder overview. Specified dimensions are related to the output of the corresponding layer. CB: Convolution Block, DSCB: Depthwise Separable Convolution Block. }
    \label{fig:encoder-overview}
\end{figure*}

As shown on Figure \ref{fig:encoder-overview} the encoder is made up of a large composition of Convolution Blocks (CB) and Depthwise Separable Convolution Blocks (DSCB) in order to enlarge the context that will serve at the final decision layer. Using our setup, the receptive field is 961 pixels high and 337 pixels wide.

CB and DSCB include Diffused Mix Dropout (DMD) layers which correspond to a new dropout strategy we propose. This strategy is detailed in Section \ref{section-archi-dropout}.

A CB is a succession of two convolutional layers, followed by an Instance Normalization layer. A third convolutional layer is applied at the end of this block. Each convolution layer uses $3 \times 3$ kernels and is followed by a \funcintext{ReLU} activation function; zero padding is introduced to remove the kernel edge effect. While the first two convolutional layers have a $1 \times 1$ stride, the third one has a $1 \times 1$ stride for CB\_1, $2 \times 2$ stride for CB\_2 to CB\_4 and $2 \times 1$ stride for CB\_5 and CB\_6. This enables to divide the height by 32 and the width by 8 so as to reduce the memory requirement. DMD is applied at three possible locations which are just after the activation layers of the three convolutional layers.

DSCB differs from CB in two respects. On the one hand, the standard convolutional layers are superseded by Depthwise Separable Convolutions (DSC)  \cite{DSC}. The aim is to reduce the number of parameters while keeping the same level of performance. On the other hand, the third convolutional layer has a fixed stride of $1 \times 1$. In this way, the shape is preserved until the last layer of the encoder.

Residual connections with element-wise sum operator are used between blocks when it is possible, that is to say when the shape is unchanged. This enables to strengthen the parameters update of the first layers of the network during back propagation.

\subsection{Attention}
At this step, we have computed the features $f$. This is a static representation that preserves the two-dimensional aspect of the task: it never changes through the attention process.
The purpose of the attention module is to recursively produce text line representations, in the desired reading order, from top to bottom for example. In addition, it has to decide when to stop generating a new line representation, \textit{i.e.} to detect the end of the paragraph.

\subsubsection{Line features generation}
Given an input image with $L$ text lines, the attention module successively produces $L$ line features. In the following, the process of the $t^\mathrm{th}$ line will be refered as attention step $t$.
To produce the successive line features, we chose to use a soft attention mechanism, as proposed by Bahdanau in \cite{Bahdanau2014}, and more specifically a hybrid attention. The soft attention is a way to focus on specific frames among a one-dimensional sequence of frames. It involves the computation of attention weights $\alpha_{t, i}$ for each frame $i$ at attention step $t$. These attention weights sum to 1 and quantify the importance of each frame for the attention step $t$. We apply the attention on the vertical axis only, in order to focus on specific features row at each attention step $t$. This way, we need one weight per features row: we roughly expect one weight to be near 1, selecting the correct feature row and the others to be near 0. Each attention step $t$ is dedicated to process the text line number $t$. The corresponding line features $l_t$ can be computed as a weighted sum between the feature rows and the attention weights:
\begin{equation}
    \tens{l_t} = \sum_{i=1}^{H_f}\tens{\alpha_{t, i}} \cdot \tens{f_i}.
\end{equation}

As one can note, this weighted sum enables to provide line features $l_t$ to be a one-dimensional sequence and thus to use the standard CTC loss for each recognized text line.

The hybrid aspect means that attention weights are computed from both the content and the location output layers. In our case, the attention weights computation combines three elements: 
\begin{itemize}
    \item $f \in \mathbb{R}^{H_f \times W_f \times C_f}$: the features in which we want to select specific line features.
    \item $\alpha_{t-1} \in \mathbb{R}^{H_f}$: the previous attention weights. To know which line to select, we must know which ones have already been processed. This is the location-based part.
    \item $h_{W_f \cdot (t-1)} \in \mathbb{R}^{C_h}$: the decoder hidden state after processing the previous text line (each line features is made up of $W_f$ frames). It contains information about what has been recognized so far. This is the content-based part.
\end{itemize}

$f$, $\alpha_{t-1}$ and $h_{W_f \cdot (t-1)}$ cannot be combined directly due to dimension mismatching. Since we want to normalize the attention weights over the vertical axis, we must collapse the horizontal axis of the features. The idea is that we only need to know if there is at least one character in a features row to decide to process it as a text line. Thus, we can collapse this horizontal dimension without information loss. This collapse is carried out in two steps: we first get back to a fixed width of 100 (since inputs are of variable sizes) through AdaptiveMaxPooling. Then, a densely connected layer pushes the horizontal dimension to collapse. The remaining vertical representation is called $\tens{f'} \in \mathbb{R}^{H_f \times C_f}$.

Instead of using directly $\alpha_{t-1}$, we found that combining it with a coverage vector $c_t \in \mathbb{R}^{H_f}$ is more beneficial. $c_t$ is defined as the sum of all previous attention weights:
    \begin{equation}
            \tens{c_t} = \sum_{k=0}^{t}\tens{\alpha_{k}}.
    \end{equation}
    $c_t$ enables to keep track of all previous line positions processed. $c_t$ is clamped between 0 and 1 for stability; we only need to know whether a row has already been processed or not. Bigger values would not have sens. We combined $c_t$ and $\alpha_t$ through concatenation over the channel axis, leading to $i_t \in \mathbb{R}^{H_f \times 2}$.
Finally, we extract some context information from $i_t$ through a one-dimensional convolutional layer with $C_j=16$ filters of kernel size 15 and stride 1 with zero padding, followed by instance normalization. This outputs $j_t \in \mathbb{R}^{H_f \times C_j}$ which contains all the contextualised spatial information from past attention steps.

We can now compute the attention weights $\alpha_t$ as follows:
\begin{itemize}
    \item For each row $i$, we compute the associated multi-scale information $s_{t,i}$, gathering all the elements we have seen previously:
        \begin{equation}
        \tens{s_{t,i}} = \tanh(
        \tens{W_f} \cdot \tens{f'_{i}} + \tens{W_j} \cdot \tens{j_{t,i}} + \tens{W_h} \cdot \tens{h_{W_f \cdot (t-1)}} 
        ).
    \end{equation}
    Indeed, $s_{t,i}$ contains both local and global information: information from features and previous attention weights can be considered as local since they are position-dependant; and information from the decoder hidden state can be seen as global since it is related to all characters recognized so far. 
    $\tens{W_f}$, $\tens{W_j}$ and $\tens{W_h}$ are weights of densely connected layers unifying the number of channels of the elements to be summed to $C_u=256$ ($\tens{W_f} \in \mathbb{R}^{C_f \times C_u}$, $\tens{W_j} \in \mathbb{R}^{C_j \times C_u}$ and $\tens{W_h} \in \mathbb{R}^{C_h \times C_u}$).
    
    \item Score $e_{t,i}$ are then computed for each row $i$:
    \begin{equation}
        \tens{e_{t,i}} = \tens{W_a} \cdot \tens{s_{t,i}},
    \end{equation}
    where $\tens{W_a}$ are weights of a densely connected layer reducing the channel axis to get a single value ($\tens{W_a} \in \mathbb{R}^{C_u \times 1}$).
    
    \item Attention weights are finally computed through softmax activation:
        \begin{equation}
        \tens{\alpha_{t,i}} = \frac
        {\exp(\tens{e_{t,i}})}
        {\displaystyle \sum_{k=1}^{H_f}\exp(\tens{e_{t,k}})}.
    \end{equation}
\end{itemize}

We emphasize that, although the attention module produces a vertical focus $l_t$, the line recognizer has a broad view of the input signal due to the size of the receptive field. Therefore, it makes the method robust to inclined or non-straight lines.

\subsubsection{End-of-paragraph detection}
\label{section-stop-process}

One of the main intrinsic problem of paragraph recognition is the unknown number of text lines to be recognized. As the text recognition is processed sequentially, an end-of-paragraph detection is needed. To solve this problem, we compare three different approaches we named \emph{fixed-stop approach}, \emph{early-stop approach} and \emph{learned-stop approach}.

For each of these approaches, we used the CTC loss to align the line prediction (probabilities lattice $p_t$ of length $W_f$), with the corresponding line transcription $y_t$. It involves the addition of the null symbol (also known as blank), leading to a new alphabet $A' = A \cup \{blank\}$. This loss is defined as the negative log probability of correctly labeling the sequence:

\begin{equation}
\mathcal{L}_{\mathrm{CTC}}(p_t, y_t) = -\ln{p(y_t|p_t)}.
\end{equation}
$p(y_t|p_t)$ corresponds to the sum of probabilities of all the paths $\pi$ leading to the target sequence $y_t$ through the reverse operations defined by automaton $\mathcal{B}$:
\begin{equation}
p(y_t|p_t) = \sum_{\pi \in \mathcal{B}^{-1}(y_t)}p(\pi|p_t).
\end{equation}
$\mathcal{B}$ is the automaton that removes the identical successive tokens from a sequence, and then also removes the null symbols from it. Finally, the probability $p(\pi|p_t)$ can be computed as:
\begin{equation}
    p(\pi|p_t) = \prod_{i=1}^{W_f}p_{t_{\pi_i}}^i, \forall \pi \in A'^{W_f},
\end{equation}
where $p^i_{t_{\pi_i}}$ is the probability of observing label $\pi_i$ at position $i$ in the line prediction sequence $p_t$.

We also defined a stopping criterion at evaluation time to avoid infinite loops. It consists in a constant $l_\mathrm{max}$ large enough to cover the biggest paragraph of the dataset. This number can easily be set to match the datasets involved; in our case $l_\mathrm{max}=30$.

The \emph{fixed-stop approach} is the simplest way to handle the end-of-paragraph detection issue. The model iterates $l_\mathrm{max}$ times and stops, as proposed in \cite{Bluche2016}.  Additional fictive lines $[y_{L+1}, ..., y_{l_\mathrm{max}}]$ are added to the ground truth as empty strings. The idea is that the extra iterations will focus on interlines, only predicting null symbols, to avoid recognizing the same line multiple times.
During training the loss is defined as follows:
\begin{equation}
    \mathcal{L}_{\mathrm{fs}} = \displaystyle \sum_{k=1}^{l_\mathrm{max}} \mathcal{L}_{\mathrm{CTC}}(p_k, y_k).
\end{equation}
This approach leads to an additional processing cost during training and evaluation due to the extra iterations needed. 

To alleviate this issue, we propose the \emph{early-stop approach}. The idea is that we can consider that the lines have all been predicted as soon as the current prediction is an empty line (only null symbols predicted). This way, we only have to add one additional line to the ground truth: $y_{L+1}$. The loss is then defined this way:
\begin{equation}
    \mathcal{L}_{\mathrm{es}} = \displaystyle \sum_{k=1}^{L+1} \mathcal{L}_{\mathrm{CTC}}(p_k, y_k).
\end{equation}

Finally, we propose the \emph{learned-stop approach} as a more elegant way to solve this problem. It consists in learning when to stop recognizing a new text line \textit{i.e.} to detect when the whole paragraph has been processed. This end-of-paragraph detection is performed at each iteration, computing the probability $d_t$ to stop or to continue the recognition. More specifically, $d_t$ determines whether $p_t$ should be considered or not. This way, the model iterates $L+1$ times to learn to predict the end of the process. 

We decided to compute this probability from two elements: the multi-scale information $s_t$ which brings some visual information about what has already been processed and what remains to be decoded, and the decoder hidden state $h_{W_f \cdot t}$ which can contain information about what have already been recognized. Indeed, it is more likely to be the end of the paragraph if the last recognized character is a dot for example. 

To this end, we first have to collapse the vertical dimension of $s_t$ for dimension matching purposes. This is carried out in the following way:
\begin{itemize}
    \item A one-dimensional convolutional layer with kernel size 5, stride 1 and zero padding is applied on $s_t$.
    \item AdaptiveMaxPooling is used to reduce the height to a fixed value of 15 (since input are of variable sizes).
    \item A densely connected layer pushes the vertical dimension to collapse, leading to $k_t \in \mathbb{R}^{C_f}$.
\end{itemize}
    
Then, the produced tensor $k_t$ is combined with $h_{W_f \cdot t}$ through concatenation over the channel axis, leading to $b_t$.

Finally, the dimension is reduced to 2 in order to compute the probabilities $d_t \in \mathbb{R}^{2}$ through a densely connected layer of weights $W_d \in \mathbb{R}^{(C_f+C_h) \times 2}$:
\begin{equation}
    d_t = W_d \cdot b_t.
\end{equation}

This approach leads to the addition of a cross-entropy loss to the CTC loss, applied to the decision probabilities $d_t$. The corresponding ground truth $\delta_t$ is one-hot encoded, deduced from the line breaks. The cross-entropy loss is defined as follows:
\begin{equation}
    \mathcal{L}_{\mathrm{CE}}(d_t, \delta_t) = -\sum_{i=1}^{2}\delta_{t_i}\log{d_{t_i}}.
\end{equation}
The final loss is then:
\begin{equation}
\mathcal{L}_\mathrm{ls} = \displaystyle \sum_{k=1}^{L} \mathcal{L}_{\mathrm{CTC}}(p_k, y_k) + \lambda \displaystyle \sum_{k=1}^{L+1} \mathcal{L}_{\mathrm{CE}}(d_k, \delta_k),
\end{equation}
where $\lambda$ is set to 1.

These three approaches are compared through experiments presented in Section \ref{section-exp-stop}.

\subsection{Decoder}
The decoder aims at recognizing a sequence of characters from the current line features $l_t$, \textit{i.e.} a whole text line. To this end, we can process $l_t$ as we do in standard OCR applied to line images, since the vertical axis is already collapsed: $l_t$ is a one-dimensional sequence of features. First, we apply a single LSTM layer with $C_h=256$ cells that outputs another representation of same dimension $r_t$ which includes some context due to the recurrence over the horizontal axis. The LSTM hidden states $h_0$ are initialized with zeros for the first line; they are kept from one line to the next one to take advantage of the context at paragraph level. Then, a one-dimensional convolutional layer with kernel $1$ going from $256$ to $N+1$ channels is applied in order to produce $p_t$, the \textit{a posteriori} probabilities of each character and the CTC null symbol, for each of the $W_f$ frames. $N$ is the size of the character set. Best path decoding is used to get the final characters sequence. Successive identical characters and CTC null symbols are removed through the CTC decoding algorithm, leading to the final text. 
All the recognized text lines are concatenated with a whitespace character as separator to get the whole paragraph transcription $p_\mathrm{pg}$.

\subsection{Diffused Mix Dropout}
\label{section-archi-dropout}
Dropout is commonly used as regularization during training to avoid over-fitting. We introduce Diffused Mix Dropout, a new dropout strategy that takes advantage of the two main modes proposed in the literature, namely standard (std.) dropout \cite{Dropout} and spatial (or 2d) dropout \cite{SpatialDropout}. A Mix Dropout (MD) layer applies one of the two possible dropout modes randomly. It enables to take advantage of the two implementations in a single layer. Diffused Mixed Dropout (DMD) consists in randomly applying MD at different locations among a set of pre-selected locations. In the model, we use DMD with dropout probability of 0.5 and 0.25 respectively for the standard and the 2d modes. Both modes have equivalent probabilities to be chosen at each execution. The benefit of using this dropout strategy is discussed in Section \ref{section-exp-dropout} through experiments.

\section{Experimental conditions}
\label{section-experiment-cond}

This section is dedicated to the presentation of the experimental conditions: datasets, pre-processing, data augmentation strategy, post-processing, computed metrics and training details are described. 

\subsection{Datasets}
In this paper, we evaluate the \modelname{} on three popular handwriting datasets: RIMES \cite{RIMES}, IAM \cite{IAM} and READ 2016 \cite{READ2016}. 
\subsubsection{RIMES}
RIMES is a popular handwriting dataset composed of gray-scale images of French handwritten text produced in the context of writing mails scenarios. The images have a resolution of 300 dpi. In the official split, there are 1,500 pages for training and 100 pages for the evaluation. To be comparable with other works, we took the last 100 training images for validation, as usually done. Segmentation and transcription are provided at paragraph, line and word levels. We used the first two segmentation levels in this work.

\subsubsection{IAM}
We used the handwriting IAM dataset which is made of handwritten copy of text passages extracted from the LOB corpus. It corresponds to gray-scale images of English handwriting with a resolution of 300 dpi. This dataset provides segmentation at page, paragraph, line and word levels with their corresponding transcriptions. In this work, we used the line and paragraph levels with the commonly used but unofficial split as detailed in Table \ref{table:split}.

\subsubsection{READ 2016}
READ 2016 has been proposed in the ICFHR 2016 competition on handwritten text recognition. This dataset is composed of a subset of the Ratsprotokolle collection used in the READ project. Images are in color and represent Early Modern German handwriting. READ 2016 provides segmentation at page, paragraph and line levels. We ignored lines with null transcription in the ground truth leading to a small difference in the split compared to the official one. We removed the character "$\lnot$" from the ground truth since it is not a real character.

\begin{table}[!h]
    \caption{Datasets split in training, validation and test sets and associated number of characters in their alphabet}
    \centering
    \resizebox{\linewidth}{!}{
    \begin{tabular}{ c c c c c c}
    \hline
    Dataset & Level & Training & Validation & Test & Charset size\\ 
    \hline
    \hline    
    \multirow{2}{*}{RIMES} & Line & 10,532 & 801 & 778 & \multirow{2}{*}{100}\\
     & Paragraph & 1,400 & 100 & 100 & \\
     
    \hline  
    \multirow{2}{*}{IAM} & Line & 6,482 & 976 & 2,915 & \multirow{2}{*}{79}\\
     & Paragraph & 747 & 116 & 336 & \\
     
    \hline  
    \multirow{2}{*}{READ 2016} & Line & 8,349 & 1,040 & 1,138 & \multirow{2}{*}{89}\\
     & Paragraph & 1,584 & 179 & 197 & \\

    \hline
    \end{tabular}
    }
    \label{table:split}
\end{table}

\begin{figure*}
\begin{minipage}[c]{0.25\textwidth}
     \begin{center}
        \includegraphics[width=\textwidth,frame]{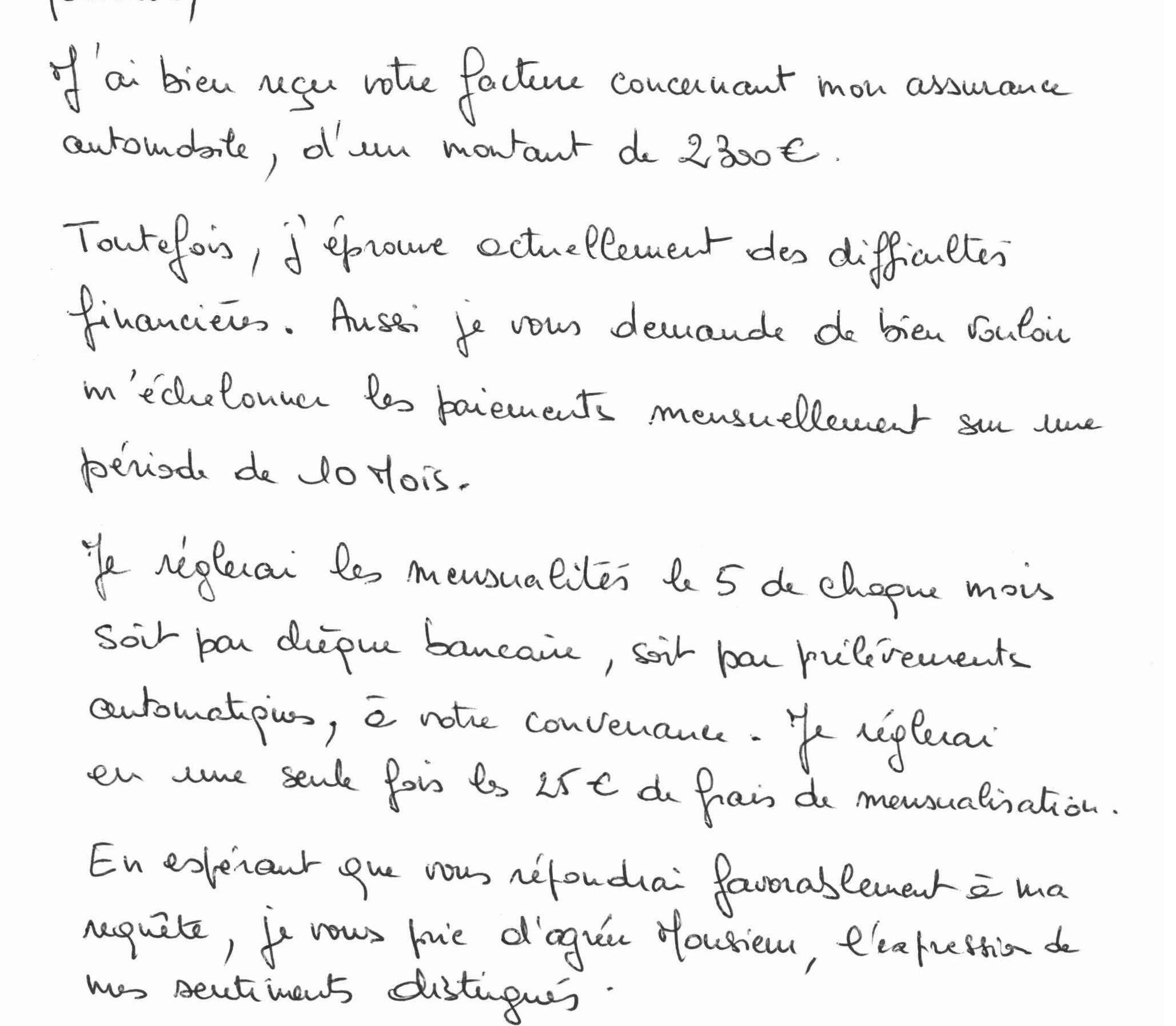}   
    \par\bigskip      
            \includegraphics[width=\textwidth,frame]{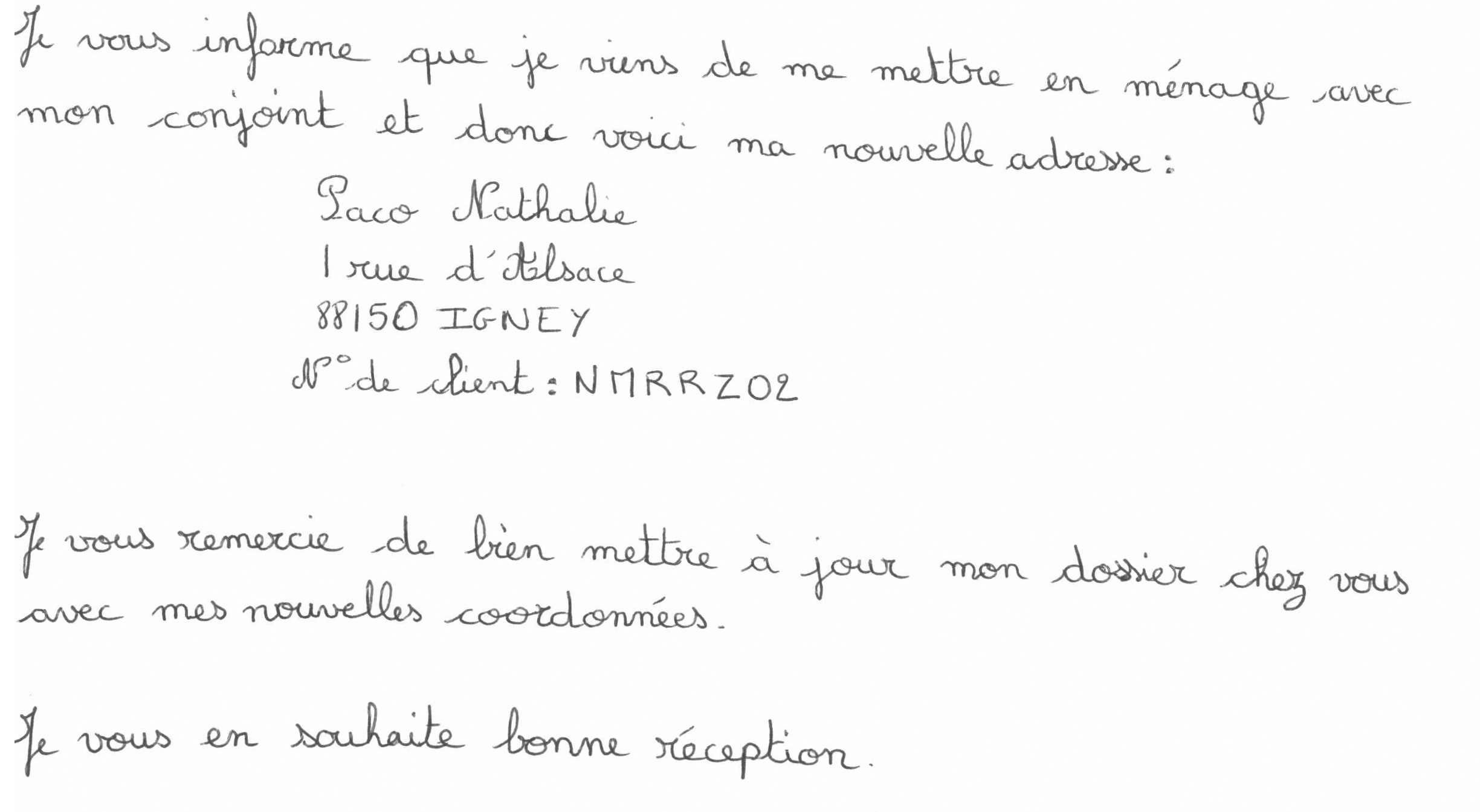}
    \par\bigskip
    \includegraphics[width=\textwidth,frame]{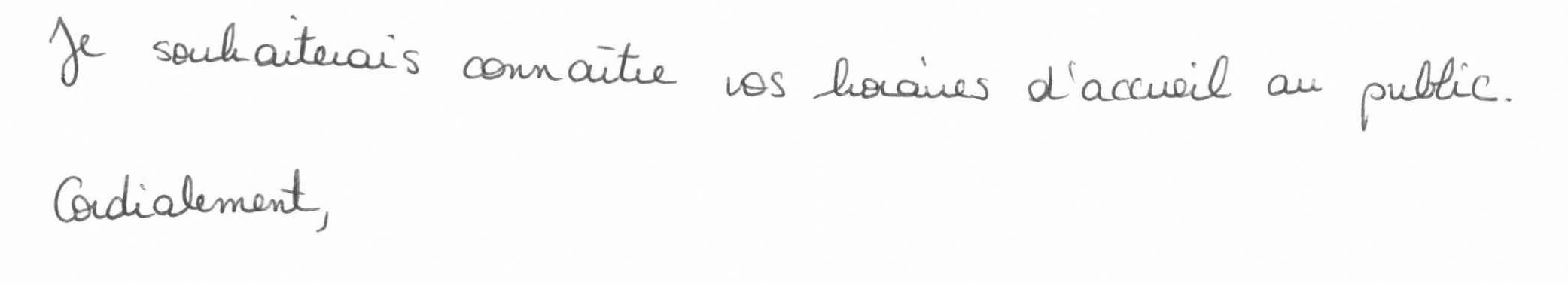}         
     \end{center}
\end{minipage}
\hfill
\begin{minipage}[c]{0.25\textwidth}
     \begin{center}
    \includegraphics[width=\textwidth,frame]{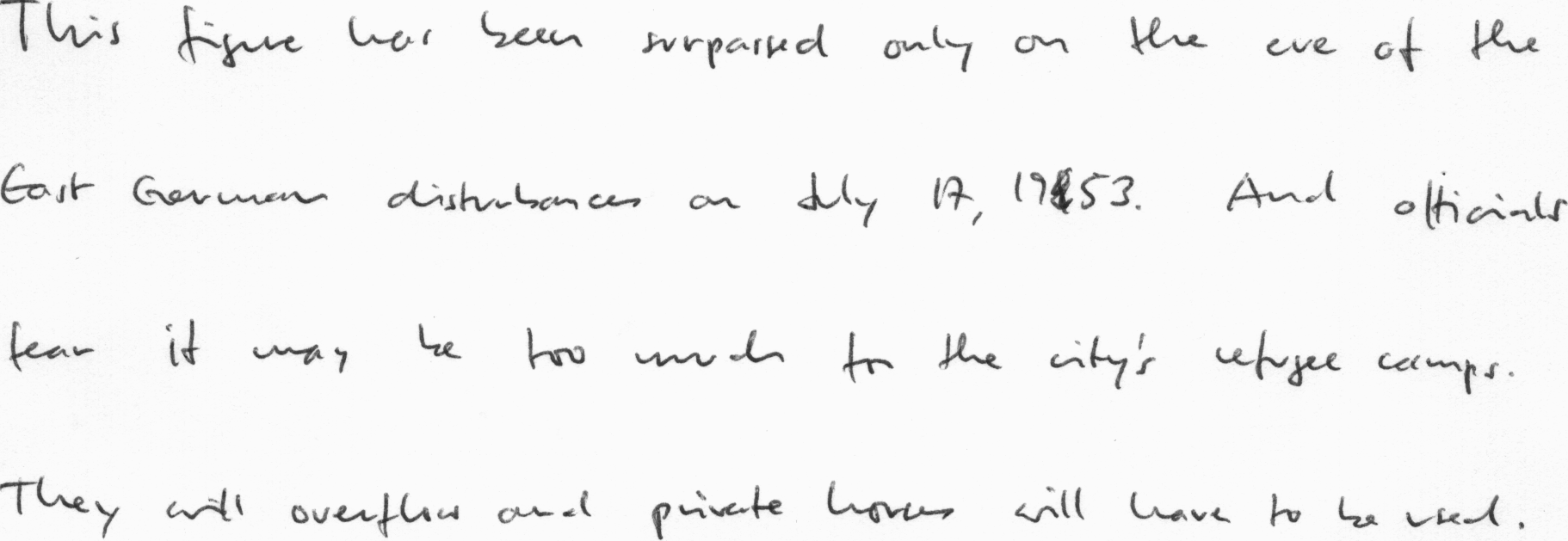}
    \par\bigskip
    \includegraphics[width=\textwidth,frame]{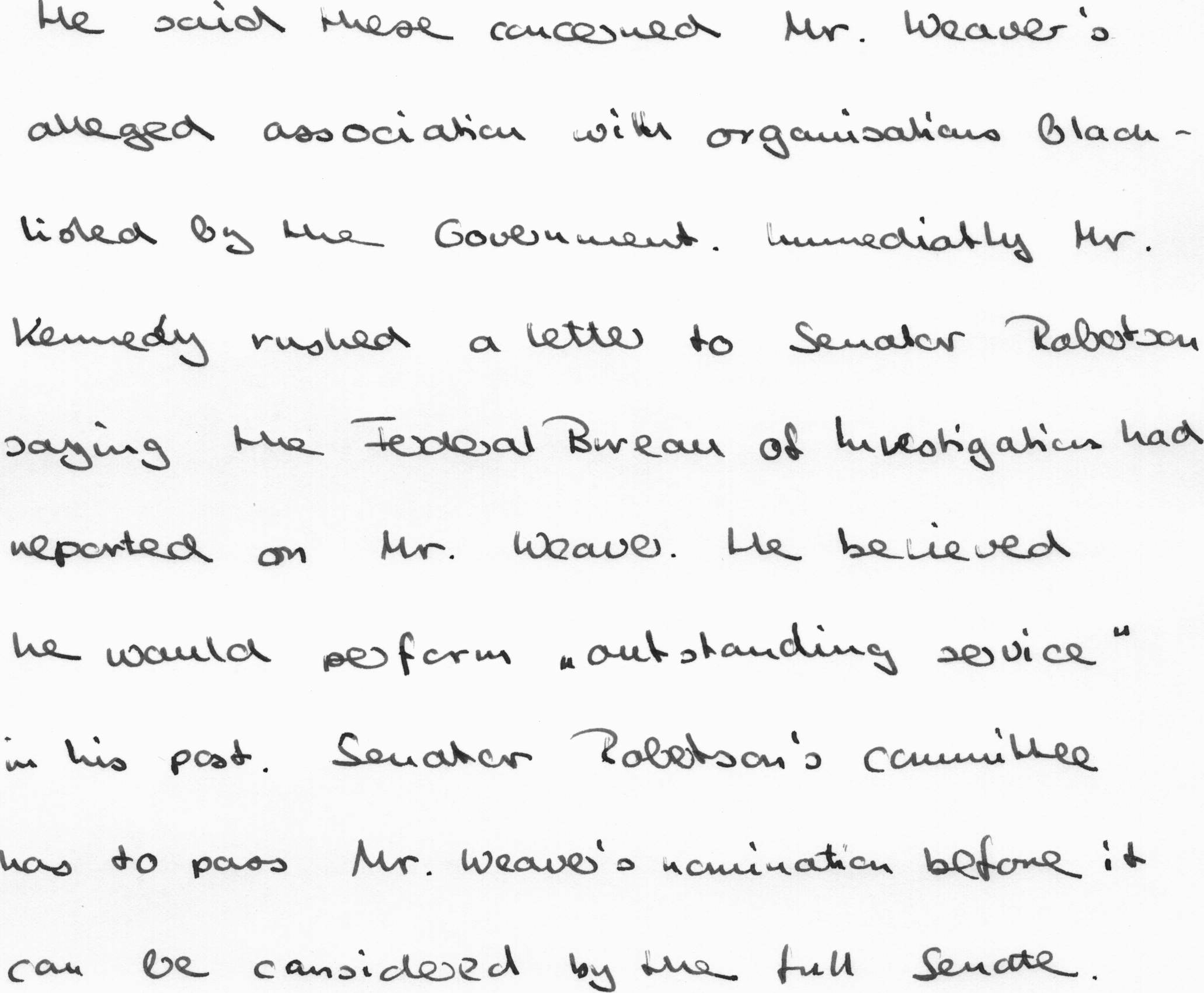}
    \par\bigskip
    \includegraphics[width=0.8\textwidth,frame]{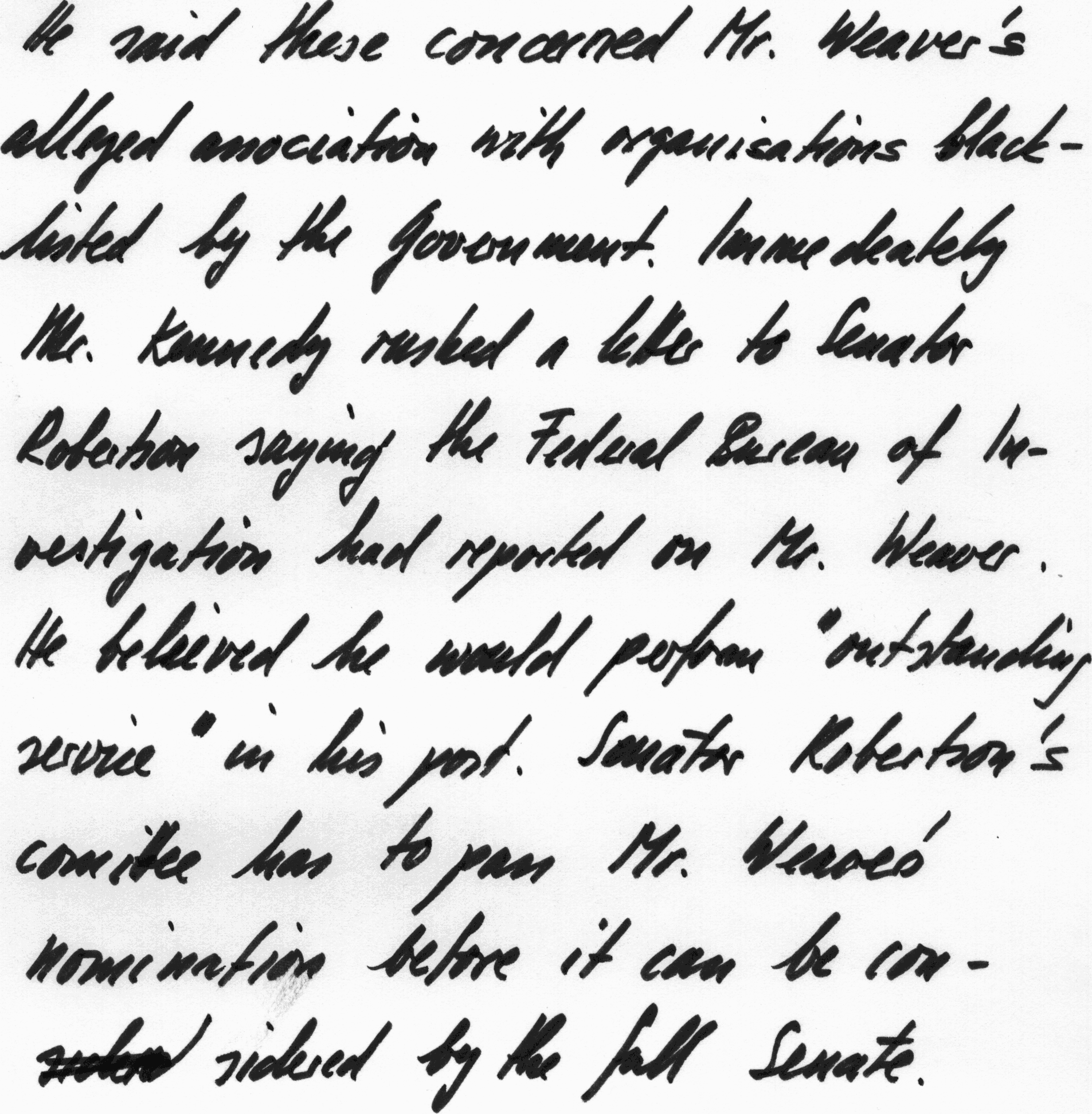}
     \end{center}
\end{minipage}
\hfill
\begin{minipage}[c]{0.25\textwidth}

     \begin{center}
     \includegraphics[width=\textwidth,frame]{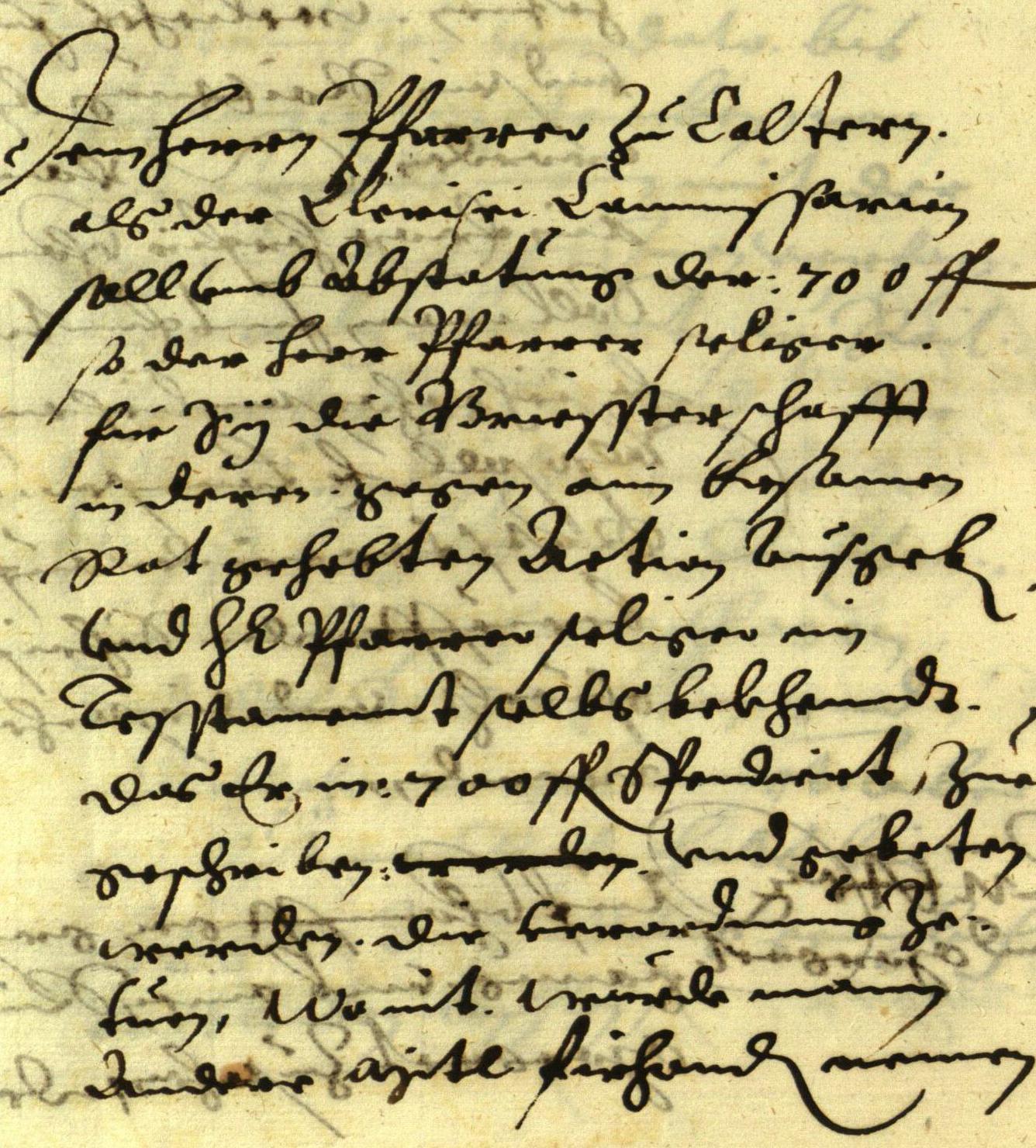}
    \par\bigskip
    \includegraphics[width=0.6\textwidth,frame]{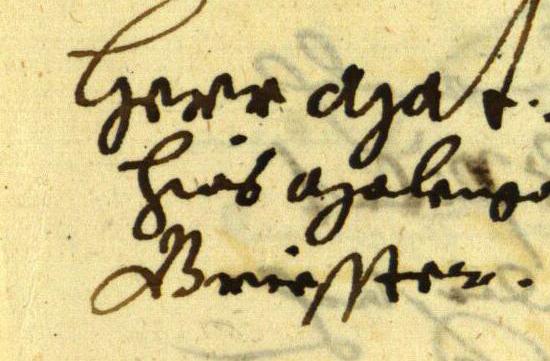}
    \par\bigskip
    \includegraphics[width=0.4\textwidth,frame]{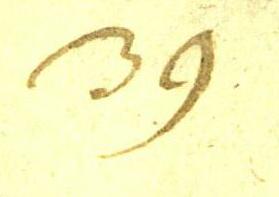}
    \end{center}
\end{minipage}

\caption{Left to right: images from the RIMES, IAM and READ 2016 datasets at paragraph level.}
    \label{fig:dataset}
\end{figure*}

Sample images from these datasets are shown in Figure \ref{fig:dataset}. As we can see, the number of lines per paragraph can vary a lot: from 2 to 18 for RIMES, from 2 to 13 for IAM and from 1 to 26 for READ 2016. RIMES exhibits more layout variability compared to the other datasets: a single paragraph image can contain multiple indents and variable interline heights. IAM layout is more structured and regular. READ 2016 sets oneself apart since its paragraph images can contain only the page number, few words or a large number of lines. In addition, these images have more noise in particular because of the bleed-through effects of handwriting occurring on the back of the pages.
In Section \ref{section-exp-pg-sota}, we show that the \modelacc{} is robust enough to handle such different datasets without adapting the model for each of them.

\subsection{Pre-processing}
We use the following pre-processings that are applied similarly on every dataset: we downscale the input images by a factor of 2 through a bilinear interpolation. We are thus working with images with a resolution of 150 dpi for IAM and RIMES for example. For the \modelacc, we zero pad the input images to reach a minimum height of 480 px and a minimum width of 800 px when necessary. This assures that the minimum features width will be 100 and the minimum features height will be 15, which is required by the model as described previously.

\subsection{Data Augmentation}
In order to reduce over-fitting and to make the model more robust to fluctuations, we set up a data augmentation strategy, applied at training time only. We used the following augmentation techniques: resolution modification, perspective transformation, elastic distortion and random projective transformation (from \cite{Yousef2020}), dilation and erosion, brightness and contrast adjustment and sign flipping. Each transformation has a probability of 0.2 to be applied. They are applied in the given order, and they can be combined except for perspective transformation, elastic distortion and random projective transformation which are mutually exclusive.

\subsection{Post-processing}
We use the well known CTC best path decoding to get the final recognition from the character probabilities lattice. We do not use any external language model or lexicon constraint. The only post-processing we used consists in keeping only one space character if several successive space characters are predicted, and removing space characters if they are predicted at the beginning or at the end of a line.

\subsection{Metrics}
In order to evaluate the quality of the recognition, we use the Character Error Rate (CER) and the Word Error Rate (WER). Both are computed with the Levenshtein distance (denoted as $\mathrm{d_\mathrm{lev}}$) between the ground truth $y$ and the recognized text $\hat{y}$, normalized by the length of the ground truth $y_\mathrm{len}$. To avoid that errors in the shortest lines have more impact on the metric than errors in the longest ones, we normalize by the total length of the ground truth:
\begin{equation}
    \mathrm{CER} = \frac{\displaystyle \sum_{i=1}^K \mathrm{d_\mathrm{lev}}(\hat{y}_i, y_i)}
                {\displaystyle\sum_{i=1}^K{y_{\mathrm{len}_i}}},
\end{equation}
where $K$ is the number of images in the data set.
WER formula is exactly the same but at word level instead of character level. We considered punctuation characters as words, as in the READ 2016 competition \cite{READ2016}.

For the evaluation of the segmentation task we present, as a comparative approach, we used two metrics: IoU and mAP. The segmentation is applied at pixel level with two classes, namely text and background. 
The IoU is defined as the intersection of text-classified pixels divided by the union of text-classified pixels between the ground truth and the prediction. We compute the global IoU over a set of images by weighting the image IoU by its number of pixels.
We compute the mAP for an image as the average of AP computed for IoU thresholds between 50\% and 95\% with a step of 5\%. Image mAPs are weighted by the number of pixels of the images to give the global mAP of a set.

Other metrics such as the number of parameters, the training time or the prediction time are useful to compare models. In the following experiments, models are trained during two days. The training time is computed as the time to reach 90\% of the convergence. This is a more relevant value since tiny fluctuations can occur after numerous epochs.

\subsection{Training details}
\label{section-training-details}
If not stated otherwise, the encoder and the last convolutional layer of the decoder of the \modelacc{} are pretrained on line-level images. It means that we transfer the weights from the architecture depicted in Figure \ref{fig:line-ocr}, which is trained on the official line-level segmented images of the same dataset. The aim of training this intermediary architecture is to only focus on the recognition aspect of the task in the first place, in order to speed up the \modelacc{} convergence and improve the recognition performance, as demonstrated in section \ref{section-pretraining}. Pretraining is carried out with the same data augmentation strategy and with the same pre-processing.

The \modelacc{} is then trained with whole paragraph images as inputs. We used the line break annotations to split the ground truth into a sequence of text line transcriptions. This enables us to use the CTC loss to train the \modelacc{} through a line-by-line alignment between the recognized text lines and the ground truth line transcriptions.

Learned-stop training and prediction processes are respectively detailed in Algorithm \ref{alg:training} and \ref{alg:prediction}. The algorithms for the fixed-stop and early-stop approaches are similar. All the instructions related to the variables $\sigma_\mathrm{CE}$, $\tens{d_t}$ and $\delta_t$ must be removed. Also, the for loop iterates $l_\mathrm{max}$ times instead of $L+1$ times for the fixed-stop approach.

\begin{algorithm}
\caption{Training process.}
\label{alg:training}
\SetKwInOut{Input}{input}
\Input{paragraph image $\tens{X}$, ground truth transcription $\tens{y}$ composed of $L$ lines $[\tens{y_1}, ..., \tens{y_L}]$}
$\tens{\alpha_0} = \tens{0}$, $\tens{h_0} = \tens{0}$, $\sigma_\mathrm{CTC}=0$, $\sigma_\mathrm{CE}=0$\tcp*[l]{Initialization}

$f = \mathrm{Encoder}(X)$\tcp*[l]{Extract 2d features $f$}
\tcp{For each line of the paragraph}
\For{t=1 \KwTo L+1}{
    \tcc{Compute attention weights $\alpha_t$ for each features row to generate 1d line features $l_t$. Also compute end-of-paragraph probabilities $d_t$}
    $\tens{l_t}$,  $\tens{\alpha_t}$ $, d_t$ $= \mathrm{Attention}(f, \tens{\alpha_{t-1}}, \tens{h_{W_f \cdot (t-1)}})$;
    
    \tcc{Determine ground truth for end-of-paragraph detection task}
    \If{$t < L+1$}
    {
        \tcc{Compute character and CTC null symbol probabilities $p_t$ for each frame of $l_t$}
        $\tens{p_t}, \tens{h_{W_f \cdot t}} = \mathrm{Decoder}(\tens{l_t}, \tens{h_{W_f \cdot (t-1)}})$\;
        
        $\sigma_\mathrm{CTC} \mathrel{{+}{=}} \mathcal{L}_\mathrm{CTC}(\tens{p_t}, \tens{y_t})$\tcp*[l]{Compute CTC loss}
        
        $\delta_t = \begin{pmatrix}
           0 \\
           1
        \end{pmatrix}$\;
    }
    \Else{
        $\delta_t = \begin{pmatrix}
           1 \\
           0
        \end{pmatrix}$\;
    }
    
    $\sigma_\mathrm{CE} \mathrel{{+}{=}} \mathcal{L}_\mathrm{CE}(\tens{d_t}, \tens{\delta_t})$\tcp*[l]{Compute cross-entropy loss}
}
backward($\sigma_\mathrm{CTC}$ $+\sigma_\mathrm{CE}$ $)$\tcp*[l]{Backpropagation}
\end{algorithm}

\begin{algorithm}
\caption{Prediction process.}
\label{alg:prediction}
\SetKwInOut{Input}{input}
\Input{paragraph image $\tens{X}$}
$l_\mathrm{max}=30$, $\tens{\alpha_0} = \tens{0}$, $\tens{h_0} = \tens{0}$, $\tens{d_t} = \begin{pmatrix}
   0 \\
   1
\end{pmatrix}$, $t=1$, $p_\mathrm{pg}=$\textquote{ } \;

$f = \mathrm{Encoder}(X)$\;
\While{$t \leq l_\mathrm{max} \And \argmax(d_t) == 1$}
{
    $\tens{l_t}$,  $\tens{\alpha_t}$ $, d_t$ $= \mathrm{Attention}(f, \tens{\alpha_{t-1}}, \tens{h_{W_f \cdot (t-1)}})$\;
    \If{$\argmax(d_t) == 1$}
    {
        $\tens{p_t}, \tens{h_{W_f \cdot t}} = \mathrm{Decoder}(\tens{l_t}, \tens{h_{W_f \cdot (t-1)}})$\;
        $p_\mathrm{pg} = \concatenate(p_\mathrm{pg}, $\textquote{ }$, \mathrm{ctc\_decoding}(\tens{p_t}))$\;
    }
    $t = t + 1$\;
}

\end{algorithm}

We can summarize those processes as follows. First, the input images are pre-processed and augmented (at training time only) as described previously. Then, the encoder extracts features $f$ from them. 
The attention module recurrently generates line features $l_t$ until $l_\mathrm{max}$ is reached or, for the learned-stop approach, until $d_t$ probabilities are in favor of stopping the process.
The decoder outputs character probabilities from each line features $l_t$, which are then decoded through the CTC algorithm, and merged together, separated by a whitespace character.
We finally get the whole paragraph transcription $p_\mathrm{pg}$.

We used the Pytorch framework with the apex package to enable mixed precision training thus reducing the memory consumption. We used the Adam optimizer for all experiments with an initial learning rate of $10^{-4}$. Trainings are performed on a single GPU Tesla V100 (32Gb). Models have been trained with mini-batch size of 16 for line-level model and mini-batch size of 8 for the segmentation model and for the \modelacc{}.

We use exactly the same hyperparameters from one dataset to another. Moreover, the model architecture is the same for each dataset: the last layer is the only difference since the datasets do not have the same character set size.
\section{Experiments}
\label{section-experiments}
\renewcommand{\thefootnote}{\alph{footnote}}

This section is dedicated to the evaluation of the \modelname{} for paragraph recognition. We show that the \modelacc{} reaches state-of-the-art results on each dataset. We study the need for pretraining on isolated text lines of the target dataset. We show that this can be avoided by using a pretrained \modelacc{} on another dataset. It enables the model to be trained on the target dataset with the paragraph-level ground truth only, without the need for line segmentation ground truth, which is a considerable practical advantage. We study the different stopping strategies on the IAM dataset. We also provide a visualization of the attention process.

\subsection{Comparison with state-of-the-art paragraph-level approaches}
\label{section-exp-pg-sota}
The results presented in this section are given for the \modelacc{}, with pretraining on line images and using the learned-stop strategy. The following comparisons are made with approaches under similar conditions, \textit{i.e.} without the use of external data (to model the language for example) and at paragraph level.

Comparative results with state-of-the-art approaches on the RIMES dataset are given in Table \ref{table:rimes-pg}. The \modelacc{} achieves better results on the test set compared to the other approaches with a CER of 1.91\% and a WER of 6.72\%. 

\begin{table}[!h]
    \caption{Recognition results  of the \modelacc{} and comparison with paragraph-level state-of-the-art approaches on the RIMES dataset.}
    \centering
    \resizebox{\linewidth}{!}{
    \begin{threeparttable}[b]
        \begin{tabular}{ l c c c c c}
        \hline
        \multirow{2}{*}{Architecture} & CER (\%) & WER (\%) & CER (\%) & WER (\%) & \multirow{2}{*}{\# Param.}\\ 
        & valid & valid & test & test\\
        \hline
        \hline
        \cite{Bluche2016} CNN+MDLSTM\tnote{a} & 2.5 & 12.0 & 2.9 & 12.6 & \\
        \cite{Wigington2018} RPN+CNN+BLSTM+LM & & & 2.1 & 9.3 & \\ 
        Ours & 1.83 & 6.26 & \textbf{1.91} & \textbf{6.72} & 2.7 M \\
        \hline
        \end{tabular}
        \begin{tablenotes}
            \item[a] With line-level attention.
        \end{tablenotes}
        \end{threeparttable}
    }
    \label{table:rimes-pg}
\end{table}

Table \ref{table:iam-pg} shows the results compared to the state of the art on the IAM dataset. As one can see, once again the \modelacc{} also achieves new state-of-the-art results with a CER of 4.45\% and a WER of 14.55\% on the test set. One can notice that we use more than 6 times fewer parameters than in \cite{Yousef2020} with 2.7 M compared to 16.4 M.

\begin{table}[!h]
    \caption{Comparison of the \modelacc{} with the state-of-the-art approaches at paragraph level on the IAM dataset.}
    \centering
    \resizebox{\linewidth}{!}{
    
    \begin{threeparttable}[b]
        \begin{tabular}{ l c c c c c}
        \hline
        \multirow{2}{*}{Architecture} & CER (\%) & WER (\%) & CER (\%) & WER (\%) & \multirow{2}{*}{\# Param.}\\ 
        & valid & valid & test & test\\
        \hline
        \hline
        \cite{Bluche2017} CNN+MDLSTM\tnote{a} & &  & 16.2 &  & \\
        \cite{Bluche2016} CNN+MDLSTM\tnote{b} & 4.9 & 17.1 & 7.9 & 24.6 & \\
        \cite{Carbonell2019} RPN+CNN+BLSTM\tnote{c} & 13.8 & & 15.6 & & \\ 
        \cite{Chung2020} RPN+CNN+BLSTM & & & 8.5 & &  \\
        \cite{Wigington2018} RPN+CNN+BLSTM+LM & & & 6.4 & 23.2 & \\
        \cite{Yousef2020} GFCN & & & 4.7 & & 16.4 M \\
        Ours & 3.02 & 10.34 & \textbf{4.45} & \textbf{14.55} & 2.7 M\\
        \hline
        \end{tabular}
        
        \begin{tablenotes}
            \item[a] With character-level attention.
            \item [b] With line-level attention.
            \item [c] Results are given for page level.
        \end{tablenotes}
        \end{threeparttable}
    }
    \label{table:iam-pg}
\end{table}

To our knowledge, there are no results reported in the literature on the READ 2016 dataset at paragraph or page level. The recognition results are presented in Table \ref{table:read2016-pg}. Our approach reaches a CER of 3.59\% and a WER of 13.94\%.
\begin{table}[!h]
    \caption{\modelacc{} results for the READ 2016 dataset at paragraph level.}
    \centering
    \resizebox{\linewidth}{!}{
    \begin{tabular}{ l c c c c c}
    \hline
    \multirow{2}{*}{Architecture} & CER (\%) & WER (\%) & CER (\%) & WER (\%) & \multirow{2}{*}{\# Param}\\ 
    & validation & validation & test & test\\
    \hline
    \hline
    Ours & 3.71 & 15.47 & \textbf{3.59} & \textbf{13.94} & 2.7 M\\
    
    \hline
    \end{tabular}
    }
    \label{table:read2016-pg}
\end{table}

The proposed \modelname{} achieves new state-of-the-art results on these three different datasets. 
For fair comparison with the other competitive approaches of the literature we highlight some comparative features in Table \ref{table:sota-details}. In this table, the approaches are analyzed with regards to the following features: 1- use of an explicit text region segmentation process 2- minimum segmentation level used (whether it is for pretraining, data augmentation or training itself) 3- number of hyperparameters adapted from one dataset to another (except for the last decision layer which is dependent of the alphabet size) 4- use of curriculum learning 5- use of data augmentation. 
Curriculum learning, which consists in progressively increasing the number of lines in the input images, can be considered as data augmentation (crop technique). It is important to note that the use of line segmentation ground truth during training is costly due to the human effort involved to create them; this is even more costly for word segmentation. Data augmentation, for its part, does not require any human effort. 

The approach proposed in \cite{Yousef2020} is the only one that does not use any segmentation label neither at line nor at word level. However, it is also the only one that requires the architecture to be adapted to each dataset. More specifically, the height and width of the input and of two intermediate upsampling layers are tuned for each dataset. It means that those 6 hyperparameters must be tuned manually to reach the performance reported, which makes the architecture not generic at all. On the contrary, the VAN architecture remains the same for every dataset in all the experiments. Another important point is that the use of line-level segmentation ground truth is not inherent to our approach, as it was only introduced as a pretraining step. As we will see in the following section, pretraining the model at line level is not mandatory. Indeed, similar performance can be obtained with paragraph-level cross-dataset pretraining, without the need for line segmentation ground truth from the target dataset, as detailed in Section \ref{section-pretraining}.

\begin{table}[!h]
    \caption{Training details of state-of-the-art approaches.}
    \centering
    \resizebox{\linewidth}{!}{
    \begin{tabular}{ l c c c c c }
    \hline
    \multirow{2}{*}{Approach} & Explicit & Min. segment. & Hyperparam. & Curriculum  & Data \\ 
    & segmentation & label used & adaptation & learning & augmentation \\
    \hline
    \cite{Carbonell2019} & \cmark & Word & \xmark & \xmark & \xmark\\
    \cite{Chung2020} & \cmark & Word & \xmark & \xmark & \cmark\\
    \cite{Wigington2018} & \cmark & Line & \xmark & \xmark & \cmark\\
    \cite{Yousef2020} & \xmark& Paragraph & 6 &\xmark  & ? \\
    \cite{Bluche2016} & \xmark & Line & \xmark & \cmark & \cmark \\
    \cite{Bluche2017} & \xmark& Line & \xmark & \cmark & \cmark \\
    Ours & \xmark & Line  & \xmark & \xmark & \cmark\\
    \hline
    \hline
    \end{tabular}
    }
    \label{table:sota-details}
\end{table}

\subsection{Impact of pretraining}
\label{section-pretraining}

In this section, we study the impact of pretraining on the \modelacc{}, using the learned-stop approach. The dataset used for pretraining is named source dataset, and the dataset on which we want to evaluate the model is named target dataset. We conducted three kinds of experiments. They are intended to evaluate the requirement of line-level segmentation labels for the target dataset to reach the best performance. 

A first training strategy consists in training the architecture at paragraph level directly on the target dataset, from scratch. 
A second strategy consists in training the architecture at paragraph level on the target dataset, with pretrained weights for the encoder and the convolutional layer of the decoder, as detailed in Section \ref{section-training-details}. In this case, pretraining is carried out on the isolated text line images of the target dataset, prior to train the \modelacc{} at paragraph level. Here, the source dataset and the target dataset are the same. This pretraining approach is referred to as line-level pretraining.
A third training strategy consists in training the architecture at paragraph level on the target dataset with weights that are initialized with those of another \modelacc{}. This other \modelacc{} is trained on a source dataset, with the second strategy. The idea is not to use any segmentation label from the target dataset. In this case the training strategy is referred to as cross-dataset pretraining.

\begin{figure}[htbp!]
\centering
\includegraphics[width=0.7\linewidth]{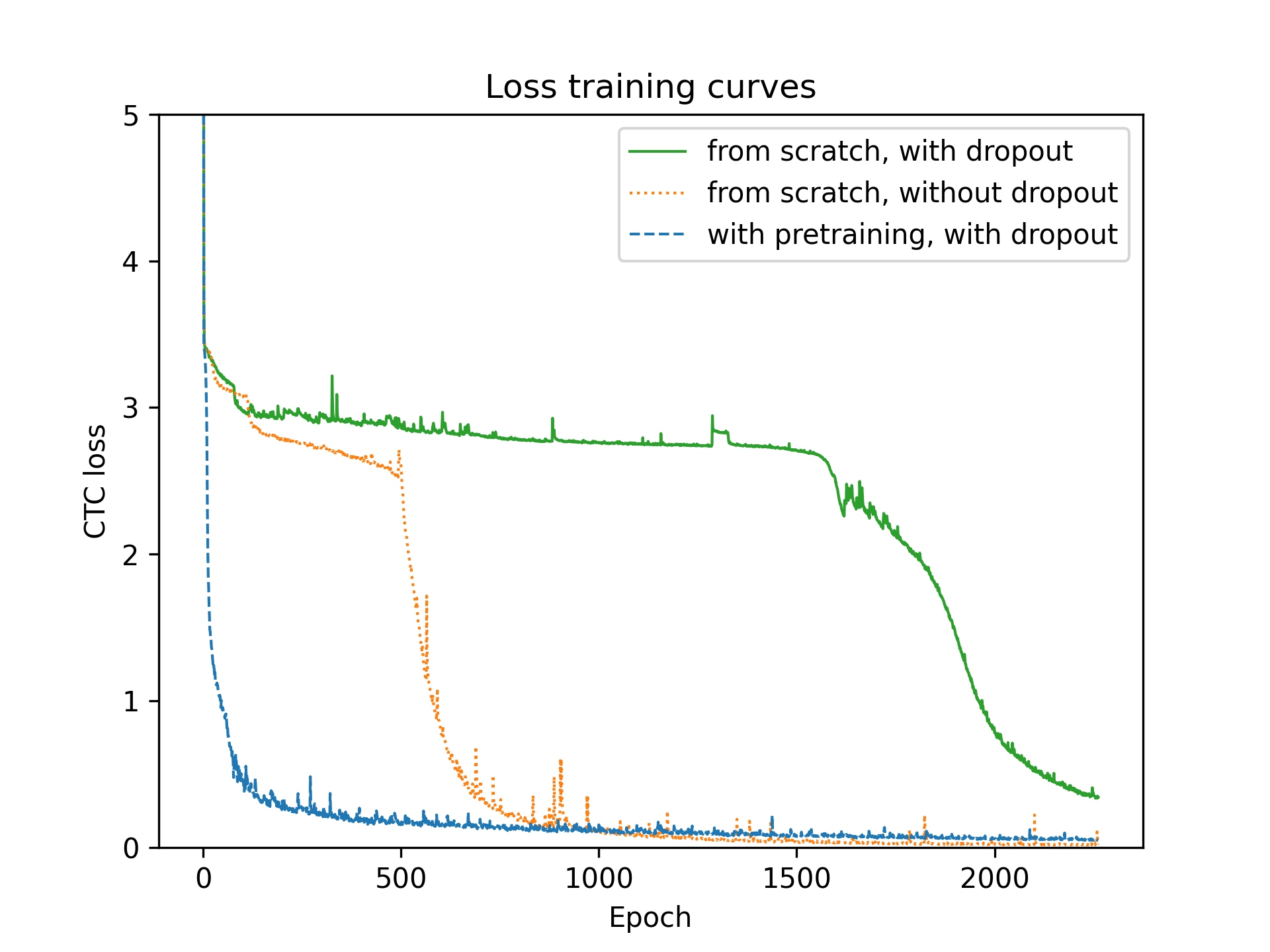}
        \caption{CTC training loss curves comparison for the \modelacc{}, with and without pretraining, on the IAM dataset.}
        \label{fig:curves-pretraining}

\end{figure}

Figure \ref{fig:curves-pretraining} shows the evolution of the CTC loss on the IAM dataset during training from scratch and training with line-level pretraining. We also highlight the impact of dropout when training from scratch. The recognition performance are given on the test set in Table \ref{table:va-pretrain}. As was expected, we can clearly notice that training the model from scratch is feasible, but it takes much time to converge and it comes at the cost of an important increase of the CER, from 4.45\% up to 7.06\%. When training from scratch, the use of dropout highly slows down the convergence but it leads to a lower CER of 7.06\% compared to 8.06\%. This experiment shows us that state-of-the art results are reached with line-level pretraining.

\begin{table}[!h]
    \caption{Impact of the pretraining on lines for the \modelacc. Results are given on the test set of the IAM dataset.}
    \centering
    \resizebox{0.7\linewidth}{!}{
    \begin{tabular}{ l c c c c }
    \hline
    Pretraining  & Dropout & CER (\%) & WER (\%) & Training time\\ 
    \hline
    \hline
    \xmark & \xmark & 8.06 & 25.38 & 1.39 d \\
    \xmark & \cmark & 7.06 & 22.65 & 1.77 d \\
    \cmark & \cmark & \textbf{4.45} & \textbf{14.55} & 0.63 d \\
    \hline
    \end{tabular}
    }
    \label{table:va-pretrain}
\end{table}

Table \ref{table:cross} reports the recognition performance of the third training strategy, using cross-dataset pretraining. In this table we report the performance of the VAN on the three datasets when pretrained on one of the two other datasets. One can notice that this pretraining strategy performs almost similarly as line-level pretraining for every datasets, but without using any segmentation label of the target dataset. This is especially true with the two similar datasets RIMES and IAM, for which the two pretraining strategies reach very similar CER: 1.97\% compared to 1.91\% for RIMES and 4.55\% compared to 4.45\% for IAM.  
Although cross-dataset pretraining is a bit less efficient on READ 2016, it still leads to competitive results. This may be explained by the differences between this dataset (RGB color encoding, historical manuscripts and language) and the other datasets (RIMES or IAM) on which pretraining is performed.

\begin{table}[!h]
    \caption{Comparison between cross-dataset pretraining and line-level pretraining for the \modelacc{}. Results are given on the test sets.}
    \centering
    \resizebox{0.7\linewidth}{!}{
    \begin{tabular}{ l c c c }
    \hline
     \multirow{2}{*}{Source dataset}& RIMES & IAM & READ 2016\\
     & CER (\%) & CER (\%) & CER (\%) \\
    \hline
    \hline
    \textbf{Cross-dataset pretraining}\\
    RIMES & \xmark & 4.55 & 4.08 \\
    IAM & 1.97 & \xmark & 4.14\\
    READ 2016 & 2.36 & 5.20 & \xmark \\
    \\
    \textbf{Line-level pretraining}\\
    Target dataset & \textbf{1.91} & \textbf{4.45} & \textbf{3.59}\\
    
    % \\
    % \textbf{From scratch}\\
    % No dropout & 3.80\\
    \hline
    \end{tabular}
    }
    \label{table:cross}
\end{table}

These experiments demonstrate the importance of pretraining for the \modelacc{}. However, we show that we can alleviate the need for line-level segmentation label of the target dataset through cross-dataset pretraining. We assume that the important pretraining effect is mainly due to the attention mechanism. Indeed, the authors of \cite{Bluche2016,Bluche2017}, who also proposed attention-based models, used curriculum learning to tackle convergence issues. We assume this is due to the direct relation between recognition and implicit segmentation (through soft attention). When the encoder is pretrained, the \modelacc{} only has to learn the attention module and the decoder, as the pretrained features may contain all the information needed for the recognition of the characters. Considering random distribution of the attention weights over the vertical axis at first, training will progressively increases values of weights where the features correspond to the correct characters. But when the encoder is randomly initialized, the features extraction must also be learned, which slows the whole training. The use of dropout makes this phenomenon even worse, but it is essential in this architecture to avoid overfitting. It seems that cross-dataset pretraining skips this issue since the attention mechanism is already learned and the encoder only needs to be fine tuned.

Finally, we can notice that without introducing any pretraining, the VAN architecture is able to converge using the paragraph annotations only, but it is not competitive anymore compared to the state of the art.
As a preliminary conclusion, we can highlight the capacity of the \modelacc{} to achieve state-of-the-art performance without no need to adapt the architecture to each dataset considered. Moreover, cross-dataset pretraining allows to reach similar results compared to line-level pretraining, without the need to use any line-level segmentation ground truth from the target dataset.

\subsection{Learning when to stop}
\label{section-exp-stop}
We now compare the three stopping strategies mentioned in Section \ref{section-stop-process}, namely fixed-stop, early-stop and learned-stop approaches.

Performance evaluation of these three methods are given in Table \ref{table:va-stop} for comparison purpose. Line-level pretraining is used for each approach, as detailed in Section \ref{section-training-details}. 
We define $d_\mathrm{mean}$, as the average of the absolute values of the differences between the actual number of lines in the image $n_\mathrm{i}$ and the number of recognized lines $n_\mathrm{r}$. This metric is used to evaluate the efficiency of the early-stop and learned-stop approaches. For $K$ images in the dataset:

\begin{equation}
    d_\mathrm{mean} = \frac{1}{K}\displaystyle \sum_{k=1}^{K} |n_{\mathrm{i}_k}-n_{\mathrm{r}_k}|.
\end{equation}

As one can see, equivalent CER and WER are obtained for each stopping strategy. Moreover, the prediction time is not significantly impacted since data formatting, tensor initialization and encoder-related computations take much longer than the recurrent process, which is made up of only a few layers at each iteration.

\begin{table}[!h]
    \caption{Comparison between fixed-stop, early-stop and learned-stop approaches with the \modelacc{} on the test set of the IAM dataset.}
    \centering
    \resizebox{\linewidth}{!}{
    \begin{tabular}{ l c c c c c}
    \hline
    Stop method  & CER (\%) & WER (\%) & Train. time & Pred. time &$d_\mathrm{mean}$\\ 
    \hline
    \hline
    Fixed & \textbf{4.41} & 14.69 & 1.20 d & 33 ms & 20.32 \\
    Early & \textbf{4.41} & \textbf{14.39} & 0.41 d & 32 ms & 0.02 \\
    Learned & 4.45 & 14.55 & 0.63 d & 32 ms & 0.03 \\
    \hline
    \end{tabular}
    }
    \label{table:va-stop}
\end{table}

\begin{figure}[htbp!]
\centering
\includegraphics[width=0.7\linewidth]{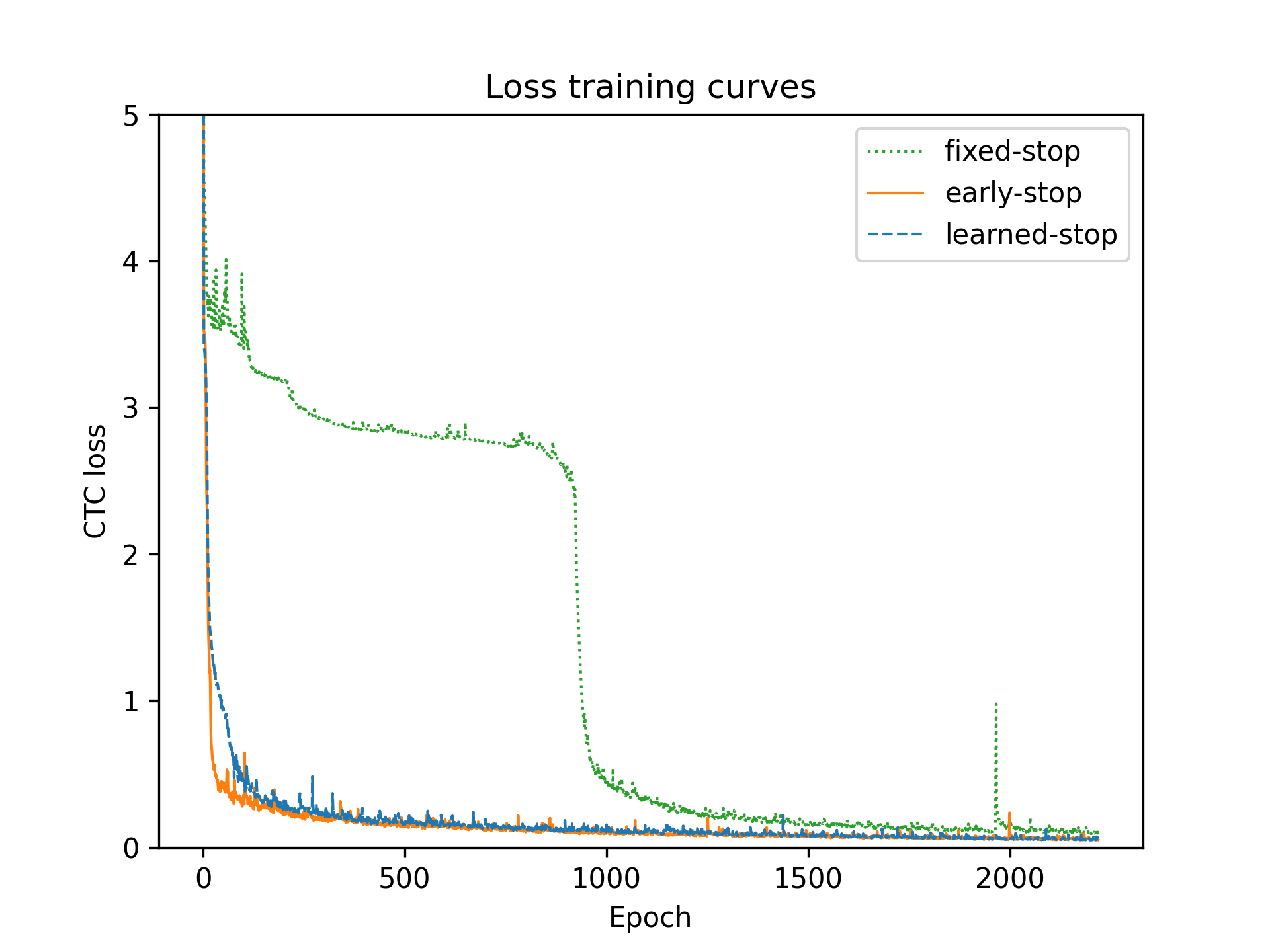}
        \caption{CTC training loss curves comparison for the \modelacc{} for each stopping approach on the IAM dataset.}
        \label{fig:loss-stop}

\end{figure}

Figure \ref{fig:loss-stop} illustrates the evolution of the CTC loss for the three approaches. One should keep in mind that the comparison is biased since the approaches do not iterate the same number of times for a same example. However, one can clearly notice that the early-stop and learned-stop approaches  converge similarly, in contrast to the fixed-stop approach, which requires far more epochs to converge. But in the end, they reach almost identical CER.

\begin{figure}[htbp!]
\centering
\includegraphics[width=0.7\linewidth]{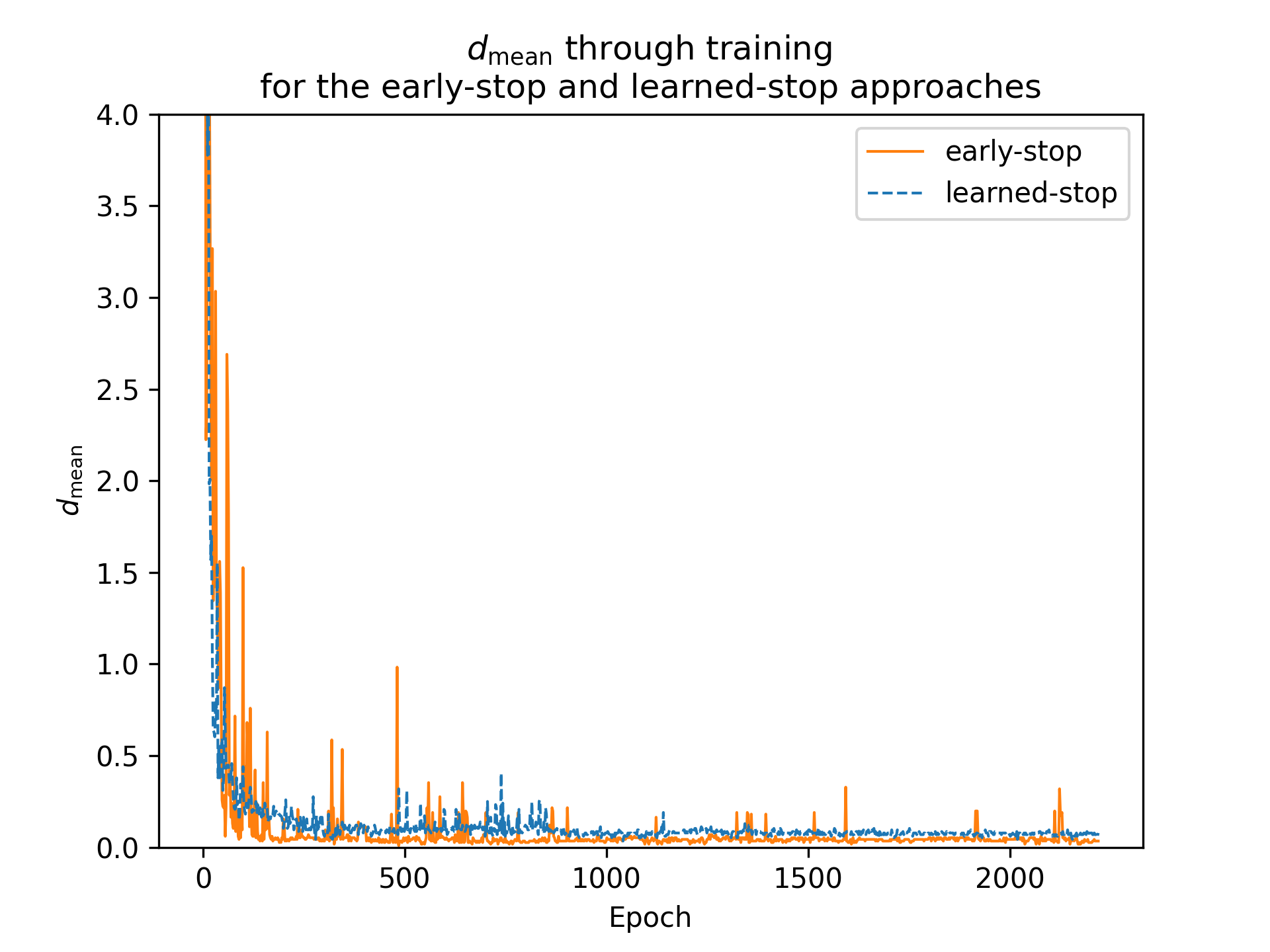}
        \caption{Comparison of the evolution of $d_\mathrm{mean}$ (the mean difference between the true and estimated number of lines in the image ) on the validation set of IAM dataset for the early-stop and learned-stop approaches.}
        \label{fig:diff-stop}
\end{figure}

Figure \ref{fig:diff-stop} compares the $d_\mathrm{mean}$ for the early-stop and learned-stop approaches on the IAM validation dataset. For visibility, the fixed-stop approach curve, which is a plateau at $d_\mathrm{mean}=20.32$, is not depicted in this figure. The learned-stop approach leads to a faster convergence of $d_\mathrm{mean}$ compared to the early-stop approach, whose curve is less stable. It results in a $d_\mathrm{mean}$ of 0.03 for the learned-stop approach and 0.02 for the early-stop approach for the test set of IAM.

With the learned-stop approach, the model successfully learns both tasks: it recognizes text lines with a state-of-the-art CER of 4.45\% and it determines when to stop with a high precision since the $d_\mathrm{mean}$ on the test set is only 0.03. It means that, on average, one line is missed or processed twice every 33 paragraph images.
We choose to keep this stopping strategy since it slightly improves the stability of the convergence through the epochs while achieving nearly identical results.

\subsection{Visualization of the vertical attention }
\begin{figure}[h!]
\centering
    \begin{subfigure}[t]{0.35\textwidth}
    \includegraphics[width=\textwidth, height=3cm, frame]{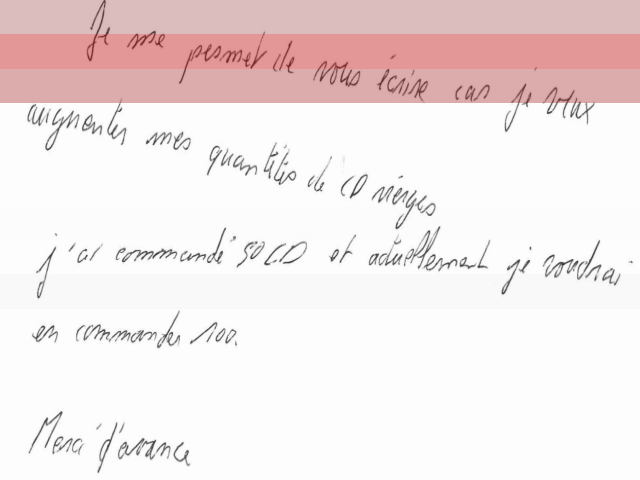}    
    \caption*{Je me permet  de vous écrire ca\textbf{s} je v\textbf{M}ux}
    \end{subfigure}

    \begin{subfigure}[t]{0.35\textwidth}
    \includegraphics[width=\textwidth, height=3cm, frame]{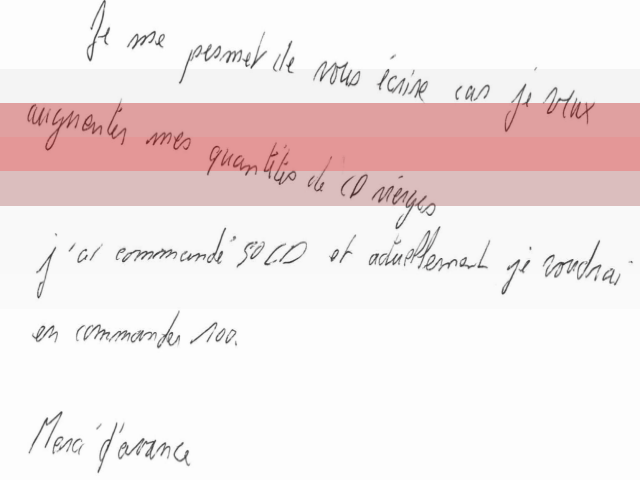}    
     \caption*{augmente\textbf{s} mes quantités de CD vierges}
    \end{subfigure}
    
    \par\bigskip

    \begin{subfigure}[t]{0.35\textwidth}
    \includegraphics[width=\textwidth, height=3cm, frame]{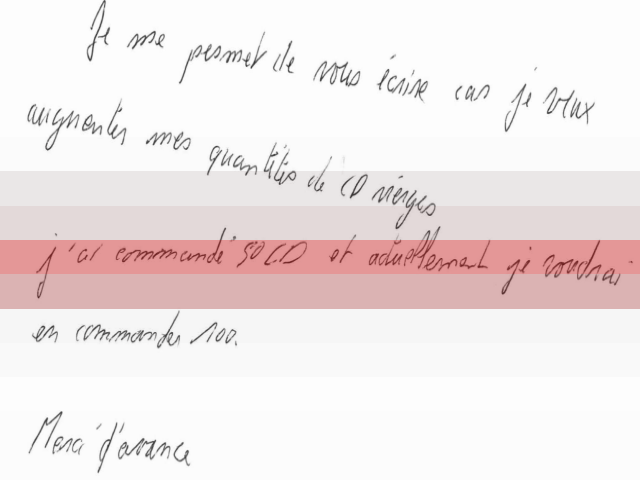}   
     \caption*{j'ai commandé 50 CD et a\textbf{d}uellement je voudrai}
    \end{subfigure}

    \begin{subfigure}[t]{0.35\textwidth}
    \includegraphics[width=\textwidth, height=3cm, frame]{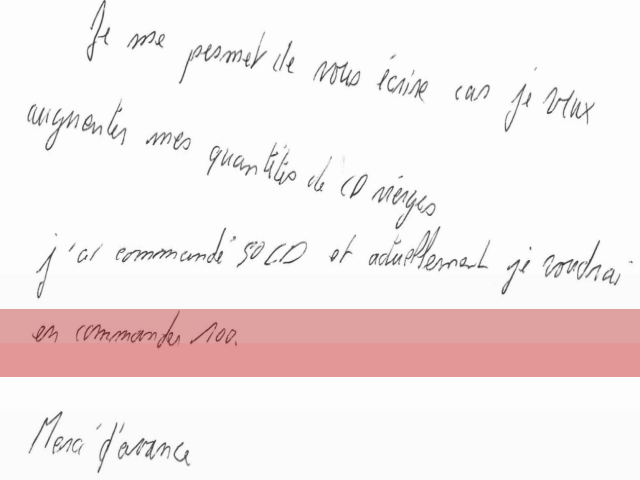}   
     \caption*{en commander 100.}
    \end{subfigure}
    
    \par\bigskip
    
    \begin{subfigure}[t]{0.35\textwidth}
    \includegraphics[width=\textwidth, height=3cm, frame]{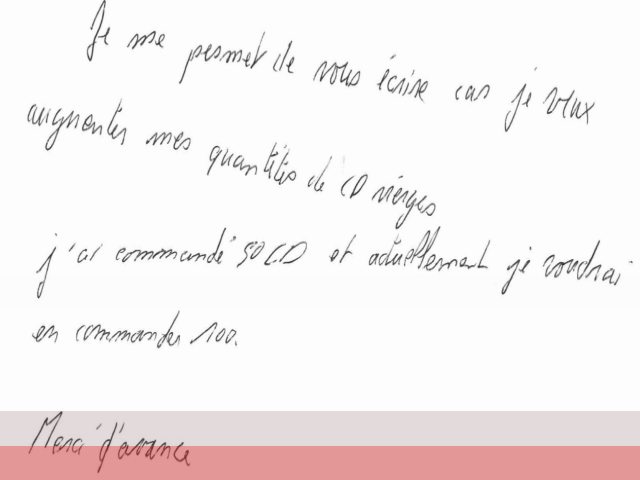}   
    \caption*{Merci d'avance}
    \end{subfigure}

    \caption{Attention weights visualization on a sample of the RIMES validation set. Recognized text is given for each line and errors are shown in bold.}
    \label{fig:viz-rimes}
\end{figure}

Figure \ref{fig:viz-rimes} shows the processing steps of an image from the RIMES validation set with a complex layout. Images from top to bottom represent the attention weights of the 5 attention iterations, each one recognizing a line. The intensity of the weights is encoded with the transparency of the red color. Given that attention weights are only computed along the vertical axis, the intensity is the same for every pixels at the same vertical position. Attention weights are rescaled to fit the original image height; indeed, attention weights are originally computed for the features height, which is 32 times smaller. The recognized text lines are given for each iteration, below the image. 

As one can see, the \modelacc{} has learned the reading order, from top to bottom. The attention weights clearly focus on text lines following this reading order. Attention weights focus mainly on one features line, with smaller weights for the adjacent vertical positions. 

One can notice that, sometimes, the focus is not perfectly centered on the text line. This may be due to rescaling, but this can also be a normal behavior due to the large size of the receptive field, which enables to manage slightly inclined lines. The second iteration shows this phenomenon very well with only one misrecognized character on an inclined line. However, processing inclined line is only possible when the lines, although inclined, do not share the same vertical position. The \modelacc{} cannot handle the case where lines would overlap vertically, because the attention weights would mix the two lines in this case.

Furthermore, one can notice that the attention is less sharp when the layout is more complex between two successive lines, as in the third image, but it does not disturb the recognition process however.

So far, we have provided strong results in favor of the \modelname{} by achieving state-of-the-art performance on three datasets. We also evaluated the need for pretraining and the efficiency of the stopping strategies we propose.

\section{Additional experimental studies}
\label{section-additional-exp}

We now provide additional results which show the superiority of the \modelacc{} on many criteria compared with a standard two-step approach (line segmentation followed by recognition).
We also highlight the positive contribution of the \modelacc{} applied to paragraph images compared with the single line recognition approaches.
Finally, we highlight the positive effect of the proposed new dropout strategy compared to standard ones.

\subsection{Comparison with the standard two-step approach}
In this section, we compare the \modelacc{} with the standard two-step approach on the IAM dataset. In this respect, we introduce two new models: a first model performs line segmentation and a second one is in charge of OCR at line level. Both models are trained separately and they do not use any pretraining strategy since they are already working at line level.

The line segmentation model follows a U-net shape architecture and is based on our FCN encoder. Indeed, the features $f$ are successively upsampled to match the features maps shapes of CB\_5, CB\_4, CB\_3, CB\_2 and CB\_1 (Figure \ref{fig:encoder-overview}). This upsampling process is handled by Upsampling Blocks (UB). UB consists in DSC layer followed by DSC Transpose layer and instance normalization. This block also includes DMD layers. Each UB output is concatenated with the feature maps from its corresponding CB. A final convolutional layer output only two feature maps, to classify each pixel of the original image between text and background.

\begin{figure*}[htbp!]
\centering
\includegraphics[width=\textwidth]{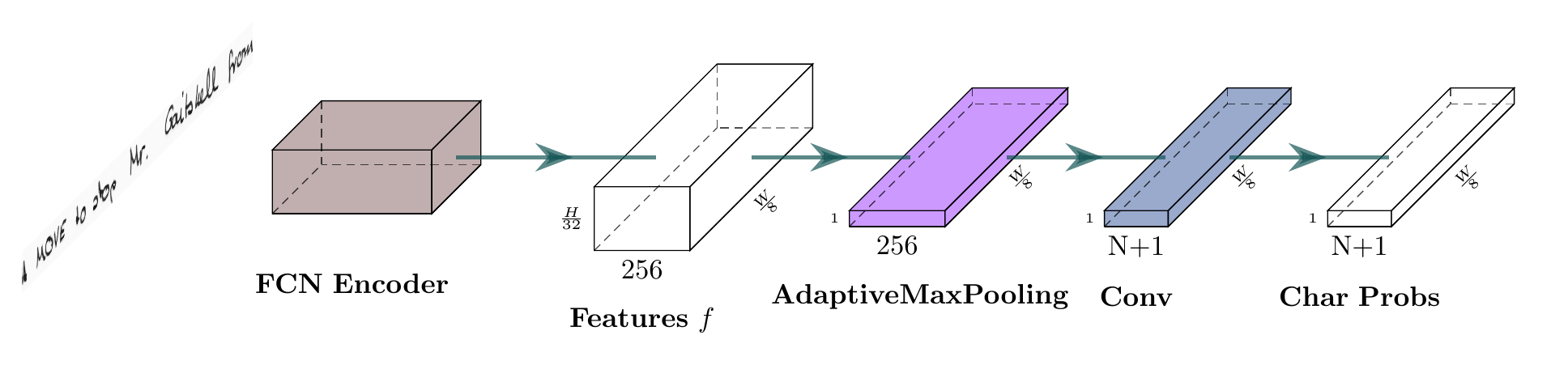}
        \caption{Text line recognition architecture overview.}
        \label{fig:line-ocr}
\end{figure*}

The OCR model for text line images is illustrated in Figure \ref{fig:line-ocr}. It is made up of the FCN encoder, followed by an AdaptiveMaxPooling layer, that pushes the vertical dimension to collapse. A final convolutional layer predicts the probability of the characters and of the CTC null symbol.

 The line segmentation model is trained on the paragraph images at pixel level with ground truth of line bounding boxes. It is trained with the cross-entropy loss. The OCR is trained on the line-level images with the CTC loss. We used a mini-batch size of 8 for the segmentation task and of 16 for the OCR.

We now detail the two steps of this approach. In a first step, paragraphs are segmented into lines: 
\begin{itemize}
    \item Ground truth bounding boxes are modified in order to avoid overlaps: we divide their height by 2.
    \item A paragraph image is given as input of the network.
    \item A 2-class pixel segmentation (text or background) is output from the network.
    \item Adjacent text-labeled pixels are grouped to form connected components.
    \item Bounding boxes are created as the smallest rectangles containing each connected component; their height is multiplied by 2.
    \item The input image is cropped using those bounding boxes to generate lines.
\end{itemize}

In a second step, the OCR model, trained on the IAM dataset at line level, is applied on those cropped line images. The segmented lines are ordered by their vertical position (from top to bottom) and the recognized text lines are concatenated to compute the CER and the WER at paragraph level.

The performances of both tasks taken separately and together are shown in Table \ref{table:seg-reco}. As one can see, the results are good for both tasks separately: we get 81.51\% for the  IoU and 85.09\% for the mAP concerning the segmentation task; and a CER of 5.01\% and a WER of 16.49\% for the OCR. 
However, when we take the output of the segmentation as input for the OCR, it leads to a CER increase of 1.54 points. Indeed, the line segmentation errors induce recognition errors.

\begin{table}[!h]
    \caption{Results of the two-step approach on the test set of IAM.}
    \centering
    \resizebox{\linewidth}{!}{
    \begin{tabular}{ l c c c c c}
    \hline
    Architecture & IoU (\%) & mAP (\%)& CER (\%) & WER (\%) & \# Param.\\ 
    \hline
    \hline
    Line seg. model & 81.51 & 85.09 & & & 1.8 M\\
    OCR on lines & & & 5.01 & 16.49 & 1.7 M \\
    Two-step approach & 81.51 & 85.09 & 6.55 & 18.54 & 1.8+1.7M\\
    \hline
    \end{tabular}
    }
    \label{table:seg-reco}
\end{table}

We can now compare the \modelname{} to the two-step approach. Comparison on the IAM test set is summarized in Table \ref{table:summary}. First, one can notice that the \modelacc{} reaches a better CER of 4.45\% compared to 6.55\%.
Prediction time is computed as the average prediction time, on the test set, to process a paragraph image. As one can see, even though the segmentation step is without recurrence, it requires much more time for prediction due to the formatting of the input required for the OCR, including the bounding boxes extraction from the original images. Moreover, it cumulates prediction times of the two models involved. 
Despite its recurrent process, the total prediction time for the \modelacc{} is shorter than that of the two-step approach since one iteration is very fast. In addition, it implies fewer parameters, this is notably due to the two models required by the two-step approach. Except for the training time, which is a bit higher for the \modelacc{}, it only provides advantages compared to the two-step approach.

\begin{table}[!h]
    \caption{Comparison of the two-step approach with the \modelname{}, results are given for the test set of the IAM dataset.}
    \centering
    \resizebox{\linewidth}{!}{
    \begin{tabular}{ l c c c c c }
    \hline
    \multirow{2}{*}{Architecture} & \multirow{2}{*}{CER (\%)} & \multirow{2}{*}{WER (\%)} & \multirow{2}{*}{\# Param.} & Training & Prediction \\ 
    & &  & & time & time \\
    \hline
    \hline
    Two-step & 6.55 & 18.54 & 1.8+1.7 M & 0.03+0.59 d & 749+28 ms\\
   VAN & \textbf{4.45} & \textbf{14.55} & 2.7 M & 0.59+0.63 d & 32 ms \\
    
    \hline
    \end{tabular}
    }
    \label{table:summary}
\end{table}

\subsection{Line level analysis}
\label{section-exp-line}

In this section, we compare the results of the \modelacc{} with the state-of-the-art approaches evaluated in similar conditions \textit{i.e.} at line level and without using external data. We also provide results for the line-level OCR model (Figure \ref{fig:line-ocr}), which has the same encoder, and for the \modelacc{} applied at paragraph level. The aim is to highlight the contribution of the \modelacc{} processing full paragraph images instead of isolated lines.

To have a fair comparison between the results obtained at paragraph and line level, one should consider the difference in the ground truth: the transcriptions of paragraph contain line breaks (or space characters in our case) between the different line transcriptions. Table \ref{table:linebreak} shows the \modelacc{} results difference at paragraph level when removing the interline characters from the metrics. As one can see, the removal of the interline character leads to an increase of the CER of 0.24, 0.09 and 0.24 on the test set of RIMES, IAM and READ 2016 respectively. The WER is not impacted by this modification. In the following tables of this section, we will use the results without considering interline characters for fair comparison with line-level approaches.

\begin{table}[!h]
    \caption{\modelacc{} results at paragraph level with and without interline characters in ground truth for RIMES, IAM and READ 2016 dataset.}
    \centering
    \resizebox{\linewidth}{!}{
    \begin{threeparttable}[b]
        \begin{tabular}{ l c c c c c }
        \hline
        \multirow{2}{*}{Dataset} & Line  & CER (\%) & WER (\%) & CER (\%) & WER (\%) \\ 
        & break & valid & valid & test & test\\
        \hline
        \hline
        \multirow{2}{*}{RIMES} & \cmark & \textbf{1.83} & \multirow{2}{*}{6.26} & \textbf{1.91} & \multirow{2}{*}{6.72} \\
        & \xmark & 2.10 &  & 2.15 & \\
        \hline
        \multirow{2}{*}{IAM} & \cmark  & \textbf{3.02} & \multirow{2}{*}{10.34} & \textbf{4.45} & \multirow{2}{*}{14.55} \\
        & \xmark & 3.07 & & 4.54 & \\
        \hline
        \multirow{2}{*}{READ 2016} & \cmark  & \textbf{3.71} & \multirow{2}{*}{15.47} & \textbf{3.59} & \multirow{2}{*}{13.94} \\
        & \xmark & 4.01 & & 3.83 & \\
        \hline
        \end{tabular}
    \end{threeparttable}
    }
    \label{table:linebreak}
\end{table}

Table \ref{table:rimes-line} shows state-of-the-art results on the RIMES dataset at line level. We report competitive results with a CER of 3.04\% for the line-level model and 3.08\% for the \modelacc{} on the test set compared to the model of \cite{Puigcerver2017} which reached 2.3\%. On should notice that Puigcerver \textit{et al.} \cite{Puigcerver2017} does not use exactly the same dataset split for training and validation. In conclusion we can highlight the performance of the VAN obtained at paragraph level which achieves a CER of 2.15\% on the test set, which corresponds to decreasing the CER by 0.93 compared to processing isolated lines.
\begin{table}[!h]
    \caption{Comparison with the state of the art on the line-level RIMES dataset.}
    \centering
    \resizebox{\linewidth}{!}{
    \begin{threeparttable}[b]
        \begin{tabular}{ l c c c c r}
        \hline
        \multirow{2}{*}{Architecture} & CER (\%) & WER (\%) & CER (\%) & WER (\%) & \multirow{2}{*}{\# Param.}\\ 
        & valid & valid & test & test\\
        \hline
        \hline
        \cite{Voigtlaender2016} 2D-LSTM+LM &  & & 2.8 & \textbf{9.6} & \\
        \cite{Puigcerver2017} CNN+BLSTM\tnote{a} & 2.2 & 9.6 & \textbf{2.3} & \textbf{9.6} & 9.6 M\\
        Ours (line-level model) & 2.20 & 6.26 & 3.04 & 8.32 & 1.7 M\\
        Ours (\modelacc{} on lines) & 1.97 & 6.09 & 3.08 & 8.14 & 2.7 M \\
        \hline
        Ours (\modelacc{} on paragraphs) & 2.10 & 6.26 & \textbf{2.15} & \textbf{6.72} & 2.7 M \\
        \hline
        \end{tabular}
        
        \begin{tablenotes}
        \item [a] This work uses a slightly different split (10,203 for training, 1,130 for validation and 778 for test).
        \end{tablenotes}
    \end{threeparttable}
    
    }
    \label{table:rimes-line}
\end{table}

Comparison with state-of-the-art results on the IAM dataset is presented in Table \ref{table:iam-line}. We reach competitive results with a CER of 4.97\% on the test set. Models proposed in \cite{Yousef_line} and \cite{Michael2019} reach similar results but the former implies a large number of parameters compared to ours and the latter is more complex including a recurrent process with attention at character level. It should be noticed however that, in \cite{Puigcerver2017,Michael2019,Yousef_line}, the authors use a slightly different split from ours. As a matter of fact, since the \modelacc{} training implies pretraining at line level, it was not possible to use the same split since some lines for training and validation are extracted from the same paragraph image for example. On the IAM dataset, the VAN also reaches a better CER at paragraph level than at line level with 4.54\% compared to 4.97\%.
\begin{table}[!h]
    \caption{Comparison with the state of the art at the line level on the IAM dataset.}
    \centering
    \resizebox{\linewidth}{!}{
    \begin{threeparttable}[b]
        \begin{tabular}{ l c c c c r}
        \hline
        \multirow{2}{*}{Architecture} & CER (\%) & WER (\%) & CER (\%) & WER (\%) & \multirow{2}{*}{\# Param.}\\ 
        & validation & validation & test & test\\
        \hline
        \hline 
        \cite{Voigtlaender2016} CNN+MDLSTM+LM & \textbf{2.4} & \textbf{7.1} & \textbf{3.5} & \textbf{9.3} & 2.6 M \\
        \cite{Puigcerver2017} CNN+BLSTM\tnote{a}  & 3.8 & 13.5 & 5.8 & 18.4 & 9.3 M\\
        \cite{Yousef_line} GFCN\tnote{a} & 3.3 & & 4.9 & & > 10 M\\
        \cite{Michael2019} Seq2seq \tiny{(CNN+BLSTM)}\tnote{a} & & & 4.87 & & \\
        Ours (line-level model) & 3.37 & 11.52 & 5.01 & 16.49 & 1.7 M\\
        Ours (\modelacc{} on lines) & 3.15 & 10.77 & 4.97 & 16.31 & 2.7 M\\
        \hline
        Ours (\modelacc{} on paragraphs) & 3.07 & 10.34 & 4.54 & 14.55 & 2.7 M\\
        \hline
        \end{tabular}
     
        \begin{tablenotes}
        \item [a] These works use a slightly different split (6,161 for training, 966 for validation and 2,915 for test).
        \end{tablenotes}
        
    \end{threeparttable}
    }       
    \label{table:iam-line}
\end{table}

The results on the READ 2016 dataset are gathered in Table \ref{table:read2016-line}. We reached state-of-the-art CER on the test set with 4.10\% compared to 4.66\% for \cite{Michael2019}. Again, the paragraph-level \modelacc{} reaches better results than the \modelacc{} applied at line level with a CER of 3.83\%.

\begin{table}[!h]
    \caption{Comparison with the state-of-the-art line-level recognizers on READ 2016 dataset.}
    \centering
    \resizebox{\linewidth}{!}{
        \begin{threeparttable}[b]
            \begin{tabular}{ l c c c c r}
            \hline
            \multirow{2}{*}{Architecture} & CER (\%) & WER (\%) & CER (\%) & WER (\%) & \multirow{2}{*}{\# Param.}\\ 
            & validation & validation & test & test\\
            \hline
            \hline
            \cite{Michael2019} Seq2seq (\tiny{CNN+BLSTM}) & & & 4.66 & & \\
            \cite{READ2016}\tnote{a}\: CNN+MDLSTM+LM & & & 4.8 & 20.9\\
            \cite{READ2016}\tnote{b}\: CNN+RNN & & & 5.1 & 21.1\\
            Ours (line-level model) & 4.49 & 18.22 & 4.25 & 17.14 & 1.7 M\\
            Ours (\modelacc{} on lines) & 4.42 & 18.17 & \textbf{4.10} & \textbf{16.29} & 2.7 M\\
            \hline
            Ours (\modelacc{} on paragraphs) & 4.01 & 15.47 & \textbf{3.83} & \textbf{13.94} & 2.7 M\\
            \hline
            \end{tabular}
            
            \begin{tablenotes}
                \item [a] results from RWTH.
                \item [b] results from BYU.
            \end{tablenotes}
        \end{threeparttable}
    }
    \label{table:read2016-line}
\end{table}

In conclusion, one can notice that the \modelacc{}, applied on isolated lines, performs at least similarly as the line-level model, for each dataset (except for RIMES with a small CER increase of 0.04 points). It also achieves state-of-the art results at line level on the READ dataset. 

The results also highlight the superiority of the \modelname{} on the RIMES, IAM and READ 2016 datasets, applied to whole paragraph images, compared to isolated lines. Multiple factors can explain this result: segmentation ground truth annotations of text lines are prone to variations from one annotator to another, bringing variability that is not present when dealing with paragraph images directly. Indeed, the model implicitly learns to segment the lines so it does not have to adapt to pre-formatted lines; it uses more context (with a large receptive field) and uses it to focus on the useful information for the recognition purpose. 

Moreover, the \modelacc{} decoder contains a LSTM layer that may have a positive impact acting as a language model without any loss of context when moving from one line to the next, when producing the output character sequence.

A key element to reach such results with a deep network is to use efficient regularization strategies. We discuss the new dropout strategy we propose in the following paragraph.

\subsection{Dropout strategy}
We use dropout to regulate the network training and thus avoid over-fitting. We defined Diffused Mix Dropout (DMD) in Section \ref{section-archi-dropout} to improve the results of the model. We carried out some experiments to highlight the contribution of DMD over commonly used standard and 2d dropout layers. Experiments are performed with the line-level model and the \modelacc{} on the IAM dataset; \modelacc{} is pretrained with weights from the corresponding line-level model. Results for the test set are shown in Table \ref{table:dropout}. The columns from left to right correspond respectively to the number of dropout layers per block (CB and DSCB), the type of dropout layer used (mix, standard or spatial), the associated dropout probabilities, the use of the diffuse option (using only one or all dropout layers per block) and the CER and WER for both models.

\label{section-exp-dropout}
\begin{table}[!h]
    \caption{Dropout strategy analysis. Results are given for the IAM test set.}
    \centering
    \resizebox{\linewidth}{!}{
    \begin{tabular}{ l c c c c c c c r}
    \hline
    & \multirow{2}{*}{\#} & \multirow{2}{*}{type} & \multirow{2}{*}{p} &\multirow{2}{*}{diffused} &  \multicolumn{2}{c}{line-level model} & \multicolumn{2}{c}{\modelacc{}} \\ 
    & & & & & CER (\%) & WER (\%) & CER (\%) & WER (\%)\\
    \hline
    \hline
    Baseline & 3 & mix & 0.5/0.25 &\cmark & \textbf{5.01} & \textbf{16.49} & 4.45 & \textbf{14.55}\\
    (1) & 3 & std. & 0.5 &\cmark & 5.24 & 17.23 & 4.46 & 14.88 \\
    (2) & 3 & 2d & 0.25 & \cmark & 5.38 & 17.64 & 4.72 & 15.63 \\
    (3) & 1 & mix & 0.5/0.25 & \xmark & 5.33 & 17.70 & \textbf{4.43} & 15.25\\
    (4) & 1 & std. & 0.5 & \xmark & 5.56 & 18.40 & 4.78 & 15.91\\
    (5) & 1 & 2d & 0.25 & \xmark & 5.70 & 18.92 & 4.93 & 16.80\\
    (6) & 3 & mix & 0.5/0.25 & \xmark & 6.76 & 21.13 & 6.32 & 19.73\\
    (7) & 3 & mix & 0.16/0.08 & \xmark & 6.71 & 20.91 & 4.64 & 15.36 \\
    (8) & 3 & mix & 0.16/0.08 & \cmark & 7.51 & 23.60 & 5.57 & 18.50\\
    
    \hline
    \end{tabular}
    }
    \label{table:dropout}
\end{table}

In (1) and (2), Mix Dropout layers are respectively replaced by standard and 2d dropout layers, preserving their corresponding dropout probability. Using Mix Dropout leads to an improvement of 0.23 points of CER compared to standard dropout and of 0.37 compared to 2d dropout for the line-level model. These improvements are lower for the \modelacc{} with 0.01 and 0.27 points.

In (3), only one Mix Dropout is used, after the first convolution of the blocks, leading to a higher CER than the baseline, with a difference of 0.34 points for the line-level model. The CER is decreased by 0.02 points for the \modelacc{} but the WER is increased by 0.70 points.
In (4) and (5), we are in the same configuration as (3) \textit{i.e.} with only one dropout layer per block. MixDropout is superseded by standard dropout in (4) and by 2d dropout in (5) resulting in an increase of the CER compared to (3). This shows the positive impact of Mix Dropout layers in another configuration.

In (6) and (7), Mix Dropout layers are set at each of the three positions \textit{i.e.} they are all used at each execution, contrary to the baseline, which uses only one dropout layer per execution. While (6) keeps the same dropout probabilities, (7) divides them by 3. In both cases, the associated CER are higher than the baseline.

Finally, in (8), we are in the same context than the baseline, but dropout probabilities are divided by 3, leading to higher CER.

We can conclude that our dropout strategy leads to a CER improvement of 0.55 points for the line-level model and of 0.33 for the \modelacc{}, when compared to (4) and (5) that do not use Mix Dropout or diffuse option.

\section{Discussion}
\label{section-discussion}

As we have seen, the \modelacc{} achieves state-of-the-art results on multiple datasets at paragraph level. However, there is one point that should be notice. Modern deep neural systems involve many training strategies (hyperparameters, optimizer, regularization strategies, pre-processings, data augmentation techniques, transfer learning, curriculum learning, and many others). This makes the comparison between architectures very difficult as some training tricks are more suited for some architectures than some others. This is why one should be convinced that the state-of-the-art results obtained in this paper are due to the whole proposition, including training strategies, and not only to the \modelacc{} architecture. However, we have provided experimental results that show the interest of the proposed training strategies of the generic \modelacc{} architecture.

We compared different stopping strategies and showed that the \modelacc{} can learn to detect the end of paragraph. This additional task has no significant impact on the performance and slightly improves the stability through training. We also compared favorably the \modelacc{} to a standard two-step approach and showed the positive impact of processing paragraph-level images compared with line-level ones, for this architecture. The new dropout strategy we propose enabled to reach even better results.

Moreover, the \modelacc{} has multiple advantages. The \modelacc{} is robust: whether it is at paragraph or line level, and no matter the dataset used, we did not adjust any hyperparameter for each dataset. The \modelacc{} takes input of variable sizes, so it could handle whole page images without any modification. As mentioned previously, the \modelacc{} can handle slightly inclined lines. However, it is limited to layouts in which there is no overlap between lines on their horizontal projection. Indeed, this case remains to be solved. A standard n-gram language model could process the outputs of the \modelacc{} architecture but its impact on the performance remains to be determined through experiments.

However, there is still room for improvement.  Notably, we showed that the \modelacc{} needs pretraining on isolated text lines of the target dataset to reach state-of-the-art results. But the need for line-level annotations is not inherent to the \modelacc{}, this is only related to this pretraining step. As a matter of fact, we demonstrate that pretraining on another dataset (cross-dataset pretraining) can alleviate this issue, even if the datasets are really different. Indeed, for the three datasets, cross-dataset pretraining leads to results similar to those from line-level pretraining, but without using any line-level annotation for the target dataset

The \modelacc{} should be considered to process single-column text document only. As a matter of fact, as it is the case for \cite{Bluche2016} and supposedly \cite{Yousef2020}, the models are designed and limited to process single-column multi-line text documents with relatively horizontal text lines. The next step would be to focus on processing images with more complex layout such as multi-column text images.

\section{Conclusion}
\label{section-conclusion}
In this paper, we proposed the \modelname{}: a novel end-to-end encoder-decoder segmentation-free architecture using hybrid attention. It handles paragraph images of variable sizes, and we showed its efficiency on the RIMES, IAM and READ 2016 datasets. Indeed, it achieves state-of-the-art results on these datasets at paragraph level. Its implicit line segmentation process enables to recognize complex layout including inclined lines. It could easily be used for recognition of whole single-column text page. 
The resulting unified model reaches a better CER than two-step approaches, for a shorter prediction time and a lighter architecture in terms of parameters. 
The proposed new dropout strategy, based on Diffused Mix Dropout layers, leads to an improvement of 0.33 points of CER.

\ifCLASSOPTIONcompsoc
  % The Computer Society usually uses the plural form
  \section*{Acknowledgments}
\else
  % regular IEEE prefers the singular form
  \section*{Acknowledgment}
\fi
The present work was performed using computing resources of CRIANN (Regional HPC Center, Normandy, France) and HPC resources from GENCI-IDRIS (Grant 2020-AD011012155). This work was financially supported by the French Defense Innovation Agency and by the Normandy region.

\begin{figure}[H]
\centering
    \includegraphics[height=1cm]{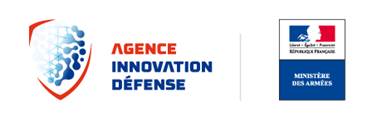}
    ~
    \includegraphics[height=1cm]{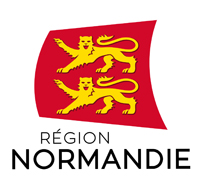}
    \label{data}
\end{figure}
\vspace{-0.7cm}

\ifCLASSOPTIONcaptionsoff
  \newpage
\fi

\bibliographystyle{IEEEtran}
\bibliography{IEEEabrv,references.bib}

\begin{IEEEbiography}[{\includegraphics[width=1in,height=1.25in,clip,keepaspectratio]{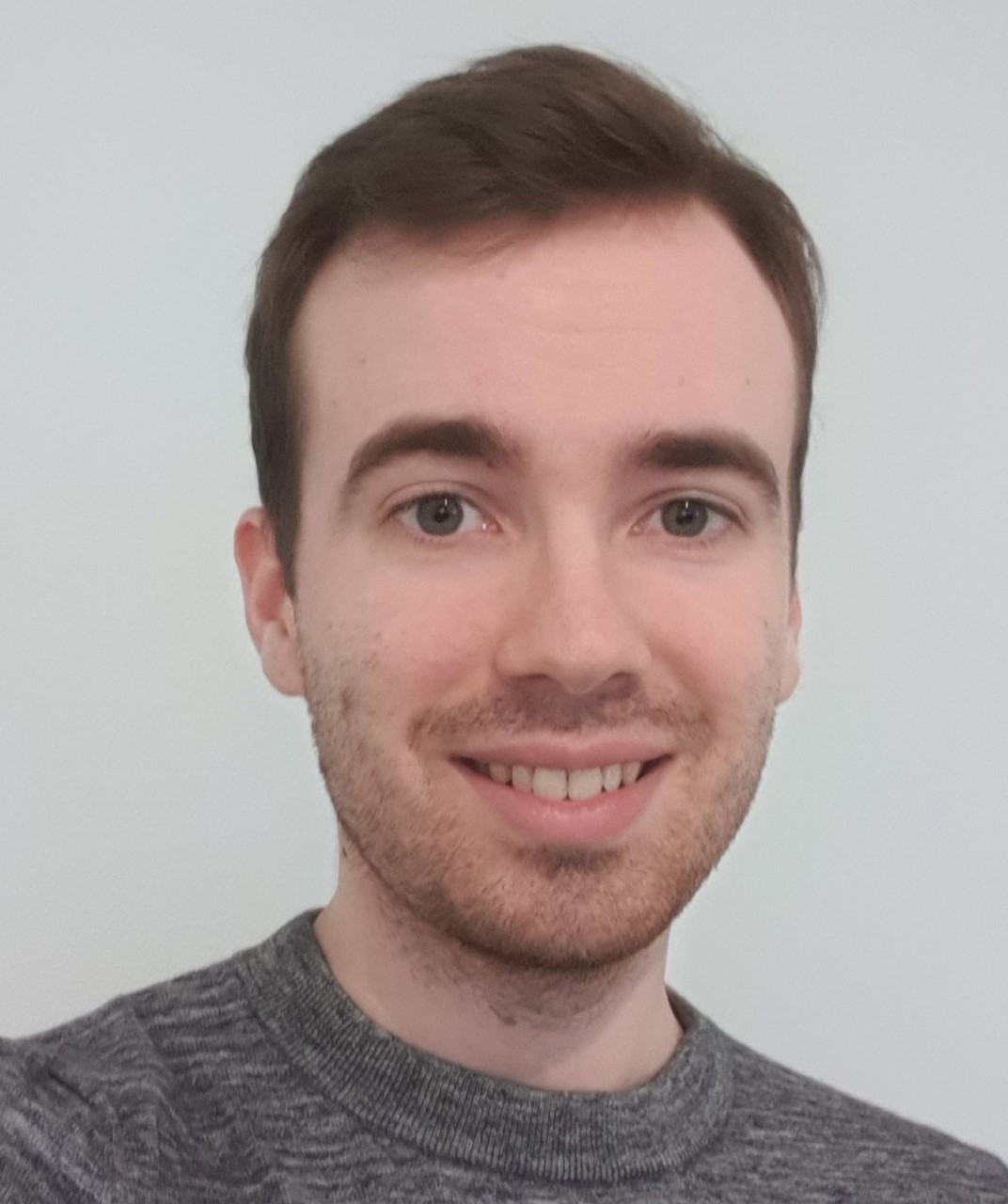}}]{Denis Coquenet}
received the Engineering Degree in Information Systems Architecture at the National Institute of Applied Sciences in Rouen, France. He is a second year Ph.D. student at the University of Rouen, France. His research interests include deep learning approaches, handwriting text recognition and more globally document analysis.
\end{IEEEbiography}

\begin{IEEEbiography}[{\includegraphics[width=1in,height=1.25in,clip,keepaspectratio]{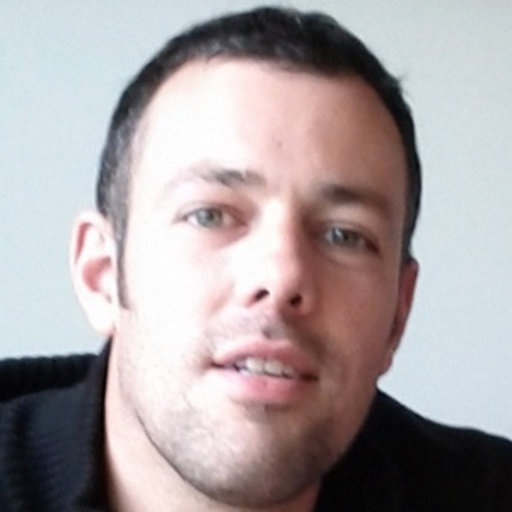}}]{Clément Chatelain}
is an Associate Professor in the Department of Information Systems Engineering at INSA Rouen Normandy, France. His research interests include machine learning applied to handwriting recognition, document image analysis and medical image analysis. His teaching interests include signal processing, deep learning and pattern recognition. In 2019, Dr. Chatelain received the French ability to conduct researches from the University of Rouen.
\end{IEEEbiography}

\begin{IEEEbiography}[{\includegraphics[width=1in,height=1.25in,clip,keepaspectratio]{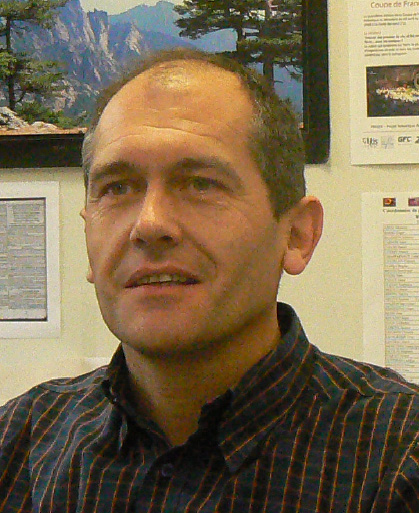}}]{Thierry Paquet}
was appointed Professor at the University of Rouen Normandie in 2002. He was the head of LITIS the research laboratory in Computer Science associated Rouen and Le Havre Universities, and Rouen INSA school of engineering. His research interests are machine learning, statistical pattern recognition, deep learning, for sequence modelling, with application to document image analysis and handwriting recognition. He has supervised 18 Ph.D students on these topics and published more than 100 papers in international conferences and scientific journals. He has contributed to many collaborative projects with academic or industrial partners. From 2008 to 2016, he was a member of the governing board of the French association for pattern recognition AFRIF. He was the president of the French association Research Group on Document Analysis and Written Communication (GRCE) from 2002 to 2010. He is regularly reviewer in main international conferences and scientific journals. From 2004 to 2012 he was in charge of the Master's degree Multimedia Information Processing at the University of Rouen Normandie.
\end{IEEEbiography}

\end{document}